\pdfoutput=1 
\documentclass[letterpaper,margin=1in]{article}

\usepackage{PRIMEarxiv}

\usepackage[utf8]{inputenc} 
\usepackage[T1]{fontenc}    
\usepackage{hyperref}       
\usepackage{url}            
\usepackage{booktabs}       
\usepackage{amsfonts}       
\usepackage{nicefrac}       
\usepackage{microtype}      
\usepackage{lipsum}
\usepackage{fancyhdr}       
\usepackage{graphicx}       
\graphicspath{{media/}}     
\usepackage{amsmath}        
\usepackage{bm}             
\usepackage{algorithm}
\usepackage{algpseudocode}
\usepackage{booktabs}
\usepackage{subfigure}
\usepackage{multirow}
\newcommand{\testtt}{\texttt}
\usepackage{amsmath} 
\usepackage{bm}      
\usepackage[numbers,sort&compress]{natbib}
\title{Modeling Parkinson's Disease Progression Using Longitudinal Voice Biomarkers: A Comparative Study of Statistical and Neural Mixed-Effects Models
}

\author{
  Ran Tong \\
  Department of Mathematical Sciences \\
  University of Texas at Dallas \\
  Richardson, TX 75080 \\
  \texttt{} \\
  \And
    Lanruo Wang \\
  Naveen Jindal School of Management \\
  University of Texas at Dallas \\
  Richardson, TX 75080 \\
  \texttt{} \\
  \And
  Tong Wang \\
  Department of Biology \\
Duke University\\
 Durham, NC 27708 \\
  \texttt{} \\
  \And
  Wei Yan\\
  Department of Neurosurgery \\
 The Second Affiliated Hospital, Zhejiang University School of Medicine\\
 Hangzhou, China\\
  \texttt{}
 }

\fancypagestyle{firstpage}{
  \fancyhf{}

  \fancyfoot[C]{%
    \scriptsize
    \parbox{0.92\textwidth}{\centering
    Published version of record: \emph{Computer Methods and Programs in Biomedicine Update},
    \href{https://doi.org/10.1016/j.cmpbup.2026.100242}{10.1016/j.cmpbup.2026.100242}.
    Version note:
    \href{https://doi.org/10.5281/zenodo.19804672}{10.5281/zenodo.19804672}.
    }
  }
}
\begin{document}
\maketitle
\thispagestyle{firstpage}

\begin{abstract}
Longitudinal voice biomarkers provide a non-invasive source of information for monitoring Parkinson's disease progression, but their statistical analysis is difficult because repeated measurements from the same subject are correlated, clinical cohorts are often small, and disease trajectories can vary substantially across individuals. This study evaluates statistical and neural mixed-effects approaches for modeling Parkinson's disease progression from telemonitoring voice data. Using the Oxford Parkinson's telemonitoring dataset ($N=42$), we compare Neural Mixed Effects (NME) models, Generalized Neural Network Mixed Models (GNMMs), and semi-parametric Generalized Additive Mixed Models (GAMMs) under the same longitudinal prediction setting. The results show that neural mixed-effects models provide flexible nonlinear representations but can overfit severely in this small-sample setting, whereas GAMMs achieve stronger predictive performance and retain interpretable smooth effects and subject-level structure. In particular, the GAMM-based approach attains the lowest prediction error (MSE 6.56), while the neural baselines have substantially larger errors (MSE $>90$). These findings support the use of interpretable statistical mixed-effects models for small longitudinal telemonitoring studies and suggest that larger and more diverse cohorts are needed before highly flexible neural mixed-effects models can be reliably assessed in this application.
\end{abstract}



\section{Introduction}

Parkinson’s disease (PD) is a progressive neuro\-degenerative disorder marked by motor symptoms such as tremor, rigidity, and postural instability, each of which lowers quality of life. The condition affects millions worldwide, and prevalence rises with age, which makes modeling the progression of Parkinson’s disease increasingly important and urgent.

In order to measure the progression of Parkinson's disease, researchers have explored several objective digital biomarkers. Wearable inertial sensors quantify gait impairment, bradykinesia, and tremor during everyday activity \cite{DelDin2016}. Smartphone accelerometers, gyroscopes, and touch-screen interactions capture movement patterns and tapping speed that relate to symptom severity \cite{Arora2015}. Handwriting and drawing tasks recorded on digitizing tablets reveal micrographia and fine-motor deficits \cite{Drotar2016}. 

Among these methods, voice biomarkers have emerged as a promising one for tracking PD progression\cite{Tsanas2012}, as it is objective, non-invasive and convenient to be obtained. Subtle shifts in pitch, loudness, and stability may appear as the disease progresses, and telemonitoring makes it possible to collect frequent longitudinal voice data that complement clinic visits.These voice features are typically used to predict scores from clinical assessments like the Unified Parkinson’s Disease Rating Scale (UPDRS) \cite{Fahn1987}, which serves as the primary response variable for quantifying symptom severity and progression in many PD studies. 

Because the same individual is measured many times, longitudinal voice data contain within‑subject correlation, and patients differ in baseline severity and rate of change. Models therefore need to represent both population trends and subject‑specific variation \cite{Mandel2021}. Furthermore, the link between high‑dimensional voice features and UPDRS may be highly nonlinear \cite{Mandel2021}.



Classical statistical methods like Linear Mixed Models (LMMs) \cite{Laird1982} and Generalized Linear Mixed Models (GLMMs) \cite{Breslow1993} have long been common tools for analyzing such longitudinal data, as they effectively use random effects to capture within-subject correlations and individual differences. However, their primary limitation lies in the inherent assumption of linear relationships for the fixed effects component, which may inadequately model the complex, nonlinear patterns of change, which are often observed in PD progression \cite{Mandel2021}.

While Nonlinear Mixed Effects (NLME) models \cite{Lindstrom1990} provide greater flexibility for nonlinear trends, their traditional optimization algorithms can be computationally demanding and may not scale efficiently to the high-dimensional parameter spaces characteristic of modern machine learning approaches \cite{Wortwein2023}. Similarly, semi-parametric extensions like Generalized Additive Mixed Models (GAMMs) \cite{Lin1999,wood2017generalized}, which use smoothing splines for time trends, can also face challenges with intricate interactions among predictors.

Deep neural networks (DNNs) offer a powerful alternative for modeling complex, nonlinear structures within large datasets. However, standard DNNs typically presuppose that observations are independent. Applying them naively to longitudinal data by disregarding these inherent correlations can result in biased estimates and suboptimal predictive performance \cite{Mandel2021}. Early adaptations, such as incorporating subject identifiers as input features in ANNs \cite{Maity2013}, aimed to address this but often encountered scalability problems as the parameter space increased with the number of subjects \cite{Mandel2021}.

More recent advancements have focused on combing neural networks with mixed-effects frameworks. A prevalent strategy has been the development of Neural Networks with Linear Mixed Effects (NN-LME), where a neural network learns nonlinear data representations, and a linear mixed model is subsequently applied to these features or forms the final layer \cite{Xiong2019}. Although these NN-LME models can capture nonlinear population-level trends, they frequently restrict person-specific (random) effects to be linear and may inherit the scalability constraints of conventional LME optimization methods \cite{Wortwein2023}.

To overcome these limitations, models such as the Generalized Neural Network Mixed Model (GNMM) \cite{Mandel2021} were introduced. The GNMM replaces the linear fixed-effect component of a GLMM with a neural network, thereby enhancing the ability to capture nonlinear associations while retaining the GLMM structure for random effects. Further extending this method, the Neural Mixed Effects (NME) model \cite{Wortwein2023} allows nonlinear person-specific parameters to be optimized at any point within the neural network architecture, offering more flexibility and scalability for modeling individual-specific nonlinear trends.

In this study, we bridge the gap between deep learning and clinical statistics by presenting, to the best of our knowledge, the first application of the Neural Mixed Effects (NME) framework \cite{Wortwein2023} to Parkinson’s Disease telemonitoring. Unlike prior studies that often treat voice recordings as independent samples or rely solely on linear random effects, we implement a fully non-linear mixed-effects architecture designed to disentangle complex subject-specific disease trajectories from population-level trends. To rigorously evaluate this approach, we benchmark the NME framework against statistical baselines, addressing a critical gap regarding the trade-off between neural flexibility and statistical stability in the small-sample regime ($N<50$) typical of clinical pilot studies.

Our contributions are threefold: 
(1) \textbf{Methodological Integration:} We introduce a novel modeling pipeline that integrates the NME \cite{Wortwein2023} and Generalized Neural Network Mixed Model (GNMM) \cite{Mandel2021} architectures into PD telemonitoring. This approach allows for the non-linear optimization of subject-specific parameters, moving beyond the linear random effects typically used in prior studies.
(2) \textbf{Rigorous Benchmarking:} We perform a comprehensive benchmark against semi-parametric statistical standards (GAMMs). We provide empirical evidence that for typical clinical trial cohort sizes ($N=42$), the high parameter space of deep neural mixed models leads to overfitting, despite their theoretical advantages.
(3) \textbf{Clinical Guidance:} We provide strategic guidance for biomedical model selection, demonstrating that while NME offers a powerful framework for large-scale data, Generalized Additive Mixed Models remain the superior choice for interpretability, stability, and predictive accuracy in current telemonitoring regimes.

\section{Related Work}

Even though some studies have leveraged the UCI Parkinson’s Telemonitoring dataset \cite{UCI2012Parkinsons} to model disease severity, many of them do not account for the longitudinal structure in the data. For example, Eskidere et al. \cite{Eskidere2012} applied various linear and nonlinear regression techniques like Support Vector Machines (SVM) and Least Squares SVM (LS-SVM) to predict UPDRS scores based on acoustic features. However, their approaches treated each observation as an independent sample, neglecting the repeated measures structure of the dataset. Similarly, Nilashi et al. \cite{Nilashi2016} proposed a hybrid system combining noise removal, clustering, and prediction methods like Adaptive Neuro-Fuzzy Inference System (ANFIS) and Support Vector Regression (SVR), but there is the absence of incorporating random effects to model individual differences in progression. Moreover, the interpretability of the model can be complicated by the combination of multiple techniques to limit the clinical utility.

Beyond static regression, recent research has increasingly adopted temporal deep learning architectures to capture disease trajectories. Long Short-Term Memory (LSTM) networks have been widely deployed to predict PD severity from voice and wearable sensors, often outperforming static baselines by retaining memory of long-term dependencies \cite{Elkholy2025, Senturk2024}. More recently, Transformer-based models, such as the Temporal Fusion Transformer (TFT), have shown promise in forecasting cognitive decline in PD by leveraging self-attention mechanisms to weigh the importance of different time points \cite{Hassan2024}. However, these deep sequence models typically require massive datasets (e.g., PPMI with hundreds of subjects) to generalize effectively. In small-$N$ cohorts typical of telemonitoring pilots, they are prone to overfitting and often lack an explicit mechanism to decompose variance into population-level trends versus subject-specific deviations \cite{Mandel2021}.

Alternative non-parametric Bayesian approaches, particularly Gaussian Processes (GPs), have also been explored for modeling PD trajectories. Comparisons of Gaussian Process Regression (GPR) with other machine learning methods have demonstrated GPR's efficacy in predicting UPDRS scores from speech signals, particularly when combined with dimensionality reduction techniques like Laplacian scores \cite{Filali2023}. Further advancements using Deep Gaussian Processes have attempted to model personalized progression patterns through multi-task learning frameworks \cite{Wang2024}. While GPs offer excellent uncertainty quantification, their computational complexity typically scales cubically with the number of observations ($O(N^3)$), making them less scalable for high-frequency telemonitoring data compared to the linear scalability of neural mixed-effects approximations.

Finally, the "black box" nature of deep learning has spurred interest in Explainable AI (XAI) for Parkinson's applications. Recent studies have integrated Layer-wise Relevance Propagation (LRP) \cite{Rezaei2024} or SHapley Additive exPlanations (SHAP) \cite{Alshammari2025} to interpret complex LSTM or CNN predictions. While these post-hoc methods provide valuable insights, they differ fundamentally from mixed-effects models, which offer \textit{intrinsic} interpretability by explicitly estimating fixed effects (population trends) and random effects (individual heterogeneity) as part of the model structure.

To improve the methods, in our study, we model the longitudinal structure of the data by comparing traditional linear mixed models (LMMs) with two recent neural extensions: the GNMM and NME model. Relatedly, we have studied longitudinal clinical time series modeling in other settings and found that compact recurrent models can remain competitive with more complex transformer-based approaches, which supports the idea that strong temporal inductive bias can matter as much as model size \cite{tong2025renaissance}. By incorporating both fixed and random effects, these models are better equipped to capture both population-level trends and subject-specific variations in disease progression. However, to date, the scalability and stability of the NME framework \cite{Wortwein2023} have not been evaluated on Parkinson's telemonitoring data, where the ratio of repeated measures to subjects is high, but the total subject count is typically low. This study fills that gap by systematically comparing these neural architectures against established statistical baselines.

\section{Methodology}
\subsection{Dataset}
This study utilizes the UCI Parkinson’s Telemonitoring Dataset\cite{UCI2012Parkinsons}, which consists of longitudinal data collected from 42 patients (28 males and 14 females) diagnosed with Parkinson’s disease. These patients were in early-stage Parkinson's disease and recruited to a six-month trial of a telemonitoring device for remote symptom progression monitoring. The recordings, comprising a range of biomedical voice measurements, were automatically collected in patients’ homes. The dataset contains a total of 5,875 records, capturing repeated measures. Each record includes 22 variables including the following information:
\begin{table}[t]
\centering
\small
\begin{tabular}{llp{3.2cm}p{7.5cm}}
\toprule
\textbf{Category} & \textbf{Variable Name} & \textbf{Type} & \textbf{Description} \\
\midrule

\multirow{3}{*}{Demographics} 
    & subject    & Integer    & Unique identifier for each subject \\
    & age           & Integer    & Age of the subject \\
    & sex           & Binary     & Subject sex (0 = male, 1 = female) \\

\midrule

\multirow{3}{*}{Clinical Scores} 
    & total\_UPDRS  & Continuous & Total UPDRS score, linearly interpolated \\
    & motor\_UPDRS  & Continuous & Motor UPDRS score, linearly interpolated \\
    & test\_time     & Continuous & Time since recruitment (in days) \\

\midrule

\multirow{16}{*}{Voice Biomarkers} 
    & Jitter(\%)     & Continuous & \multirow{5}{7.5cm}{Several measures of variation in fundamental frequency} \\
    & Jitter(Abs)    & Continuous & \\
    & Jitter:RAP     & Continuous & \\
    & Jitter:PPQ5    & Continuous & \\
    & Jitter:DDP     & Continuous & \\
\cmidrule(lr){2-4}
    & Shimmer        & Continuous & \multirow{6}{7.5cm}{Several measures of variation in amplitude} \\
    & Shimmer(dB)    & Continuous & \\
    & Shimmer:APQ3   & Continuous & \\
    & Shimmer:APQ5   & Continuous & \\
    & Shimmer:APQ11  & Continuous & \\
    & Shimmer:DDA    & Continuous & \\
\cmidrule(lr){2-4}
    & NHR            & Continuous & \multirow{2}{7.5cm}{Ratio of noise to tonal components in the voice} \\
    & HNR            & Continuous & \\
\cmidrule(lr){2-4}
    & RPDE           & Continuous & Nonlinear dynamical complexity measure \\
\cmidrule(lr){2-4}
    & DFA            & Continuous & Fractal scaling exponent \\
\cmidrule(lr){2-4}
    & PPE            & Continuous & Nonlinear measure of fundamental frequency variation \\
\bottomrule
\end{tabular}
\caption{Variable descriptions for the Parkinson’s Telemonitoring Dataset.}
\end{table}

In the early stages of Parkinson’s disease (PD), many patients show noticeable changes in their speech, such as unstable pitch, uneven loudness, hoarseness, and unclear pronunciation. Clinically, this cluster of symptoms is often referred to as \textit{hypokinetic dysarthria}. PD affects the brain's ability to control the muscles which are used for speaking, including those in the throat and chest. These impairments are largely driven by underlying \textit{bradykinesia} (slowness of movement) and muscular \textit{rigidity}. Since telemedicine and remote health tools become more common, voice recordings have become a useful and non-invasive way to monitor how the disease changes over time. The early diagnosis can also be helped. 

Many voice-based measurements are contained in the datasets and can be grouped into four types: frequency changes (Jitter), amplitude changes (Shimmer), noise features (NHR and HNR), and nonlinear patterns (RPDE, DFA, and PPE). 

Jitter features including Jitter (\%), Jitter (Abs), RAP, PPQ5, and DDP measure small changes in pitch between voice cycles. Since the vocal folds of people with PD do not move smoothly due to laryngeal rigidity and micro-tremors, they cannot keep a steady pitch which will lead to a higher jitter values.

Shimmer feature such as Shimmer in percent and decibels, APQ3, APQ5, APQ11, and DDA show how the loudness of the voice changes from one cycle to the next. Due to the muscle control problems and reduced respiratory support (hypophonia) which make their speech volume less steady, these values are resulted in higher.

NHR (Noise-to-Harmonics Ratio) and HNR (Harmonics-to-Noise Ratio) are used to check how much noise is in the voice. PD patients tend to have more noise and less clear voice sounds due to incomplete glottic closure, which make these values worse than healthy individuals.

Furthermore, Recurrence Period Density Entropy (RPDE), Detrended Fluctuation Analysis (DFA), and Pitch Period Entropy (PPE) are nonlinear dynamic measures that capture complexity and unpredictability in vocal patterns. People with PD may speak in a way which is less regular or harder to predict. These features can be helpful to identify those subtle inssues in voice control.

Overall, these voice features can help to study speech problems in Parkinson's disease. These features can be easily measured and also related to related to motor symptoms closely, therefore they are valuable digital biomarkers in tracking disease progression and supporting remote healthcare systems.

In this study, we use total\_UPDRS as the response variable to explore the impact of voice biomarkers on disease progression.

\begin{figure}[H]
    \centering
    \subfigure[Correlation heatmap]{
        \includegraphics[width=0.3\textwidth]{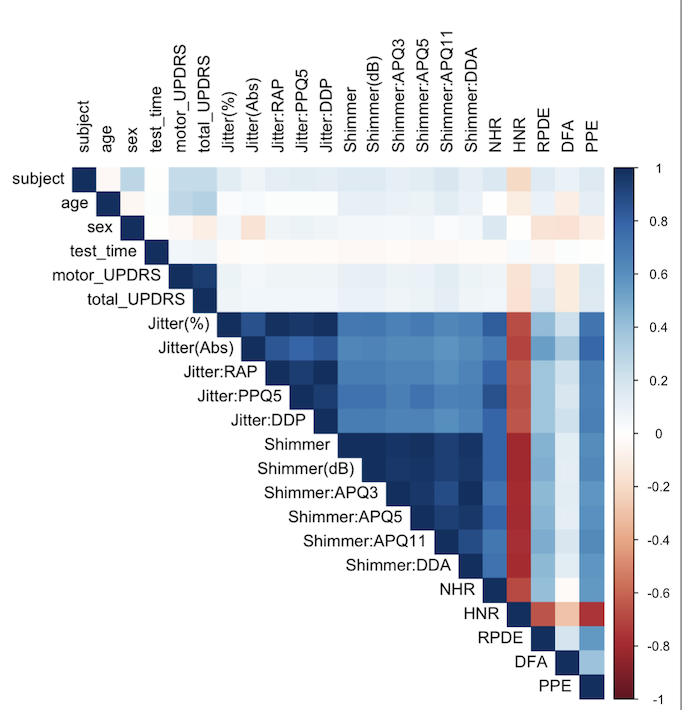}
    }
    \subfigure[Scatter plot of total\_UPDRS versus age]{
        \includegraphics[width=0.3\textwidth]{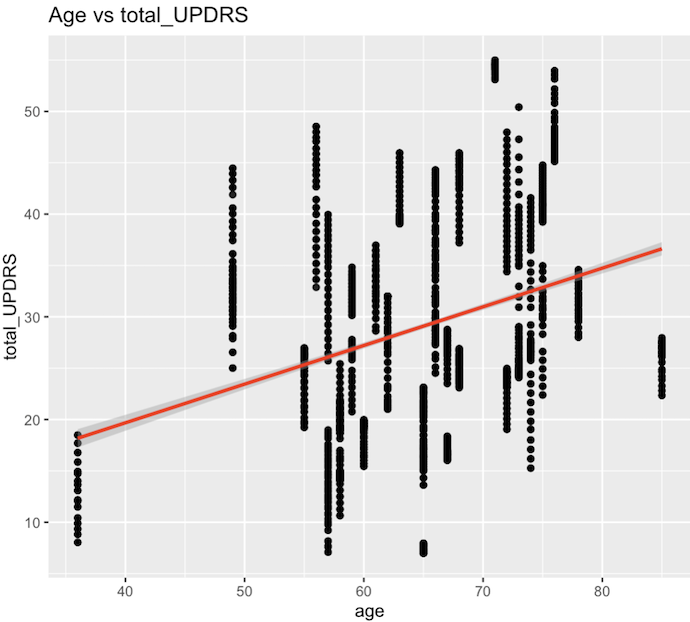}
    }
    \subfigure[Subject\-specific log(total\_UPDRS) trajectories over time]{
        \includegraphics[width=0.3\textwidth]{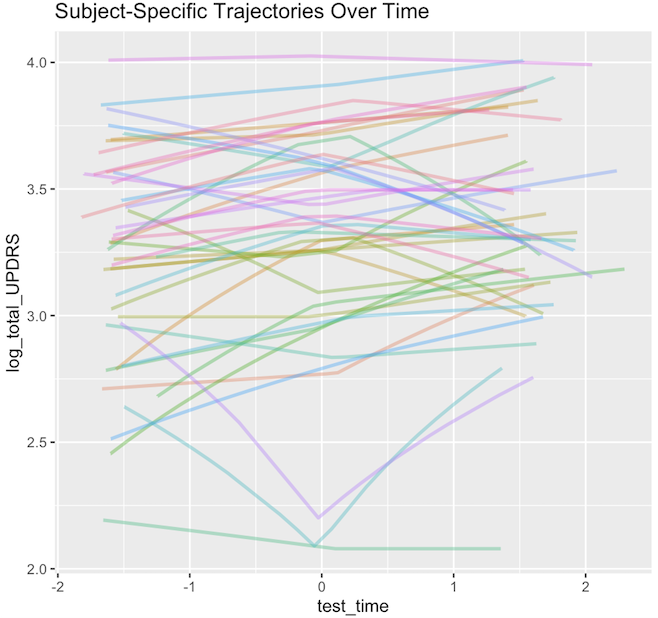}
    }
    \caption{Data analysis of the dataset}
    \label{fig:three_figs}
\end{figure}

Figure 1(a) shows the pairwise Pearson correlations among all measured variables. Several voice-based features, particularly jitter and shimmer measures, show moderate positive correlations with UPDRS scores. This indicates that voice instability is related to the severity of the disease.

Figure 1(b) illustrates the relationship between patient age and total\_UPDRS scores. Although the scatterplot reveals variability within the age range, the fitted regression line shows a clear positive association, indicating that older individuals tend to have more severe symptoms.

Figure 1(c) presents longitudinal trends in log-transformed total\_UPDRS scores for individual patients. The patterns differ among patients, with some showing progressive worsening while others remain stable or even slightly improve. The heterogeneity in progression patterns shows the necessity for individualized modeling approaches like mixed effects models.

\subsection{Traditional Methods: Linear Mixed--Effects Model (LMM)}\label{subsec:trad_LMM}
We first apply Linear Mixed-Effects model  \cite{Laird1982} on our dataset. Let $Y_{ij}$ denote the UPDRS score for subject $i\;(i=1,\dots,m)$ at time
$t_{ij}\;(j=1,\dots,n_i)$, and let
$X_{ijk}$ be the $k$‑th voice feature ($k=1,\dots,K$).
We use a linear mixed–effects model with a subject‑specific
random intercept $b_{0i}$ and random slope $b_{1i}$ for time:
\begin{equation}\label{eq:LMM_scalar}
  Y_{ij}
  \;=\;
  \beta_0
 \;+\;
  \beta_1 t_{ij}
 \;+\;
  \sum_{k=1}^K \beta_{k+1}\,X_{ijk}
 \;+\;
  b_{0i}
 \;+\;
  b_{1i} t_{ij}
 \;+\;
  \varepsilon_{ij}.
\end{equation}
In this model, the terms $\beta_0$, $\beta_1 t_{ij}$, and $\sum_{k=1}^K \beta_{k+1}\,X_{ijk}$ represent the fixed effects. Specifically, $\beta_0$ is the overall intercept, $\beta_1$ is the average slope for time $t_{ij}$ across all subjects, and $\beta_{k+1}$ are the coefficients for the $K$ voice features $X_{ijk}$, representing their average effects on $Y_{ij}$. These fixed effects describe the population-average relationships.

The terms $b_{0i}$ and $b_{1i} t_{ij}$ represent the random effects for subject $i$. Here, $b_{0i}$ is the subject-specific random intercept, showing how subject $i$'s baseline UPDRS score deviates from the overall intercept $\beta_0$. Similarly, $b_{1i}$ is the subject-specific random slope for time, indicating how subject $i$'s rate of change in UPDRS score over time $t_{ij}$ deviates from the average time slope $\beta_1$. These random effects capture individual heterogeneity around the population-average trends.

Finally, $\varepsilon_{ij}$ is the residual error term for subject $i$ at time $j$, representing within-subject variability not explained by the fixed or random effects.


\subsection*{Distributional assumptions.}
 The subject-specific random effects $(b_{0i}, b_{1i})^\top$ are assumed to be drawn from a bivariate normal distribution with a mean of zero and a covariance matrix $D$:
\[
  \begin{pmatrix} b_{0i}\\ b_{1i}\end{pmatrix}
  \,\sim\,
  \mathcal N\!\Bigl(\mathbf0,\,
     D=
     \begin{pmatrix}
       \sigma_{b0}^2 & \rho\,\sigma_{b0}\sigma_{b1}\\
       \rho\,\sigma_{b0}\sigma_{b1} & \sigma_{b1}^2
     \end{pmatrix}\Bigr).
\]
 The residual errors $\varepsilon_{ij}$ are assumed to be independent and identically distributed (i.i.d.) normal random variables with a mean of zero and variance $\sigma^2$:
\[
  \varepsilon_{ij}\sim\mathcal N(0,\sigma^2).
\]
Furthermore, the random effects $\mathbf b_i$ and residual errors $\boldsymbol\varepsilon_i$ are assumed to be independent of each other.

\subsection*{Matrix formulation.}

 Let $\mathbf y_i = (Y_{i1},\dots,Y_{in_i})^\top$ be the vector of $n_i$ UPDRS scores for subject $i$. The fixed effects design matrix $\mathbf X_i$ and the random effects design matrix $\mathbf Z_i$ for subject $i$ are defined as:
\[
  \mathbf X_i =
     \begin{pmatrix}
       1 & t_{i1} & X_{i1,1} & \dots & X_{i1,K}\\
       \vdots & \vdots & \vdots & \ddots & \vdots\\
       1 & t_{in_i} & X_{in_i,1} & \dots & X_{in_i,K}
     \end{pmatrix},
  \quad
  \mathbf Z_i =
     \begin{pmatrix}
       1 & t_{i1}\\
       \vdots & \vdots\\
       1 & t_{in_i}
     \end{pmatrix}.
\]

The matrix $\mathbf X_i$ contains a column of ones for the intercept, a column for time $t_{ij}$, and $K$ columns for the voice features $X_{ijk}$. The matrix $\mathbf Z_i$ contains a column of ones for the random intercept and a column for time $t_{ij}$ corresponding to the random slope.
Then the model for subject $i$ can be written as:
\[
  \mathbf y_i
  \;=\;
  \mathbf X_i\boldsymbol\beta
 \;+\;
  \mathbf Z_i \mathbf b_i
 \;+\;
  \boldsymbol\varepsilon_i,
  \qquad
  \mathbf b_i\sim\mathcal N(\mathbf0,D),\;
  \boldsymbol\varepsilon_i\sim\mathcal N(\mathbf0,\sigma^2I_{n_i}).
\]

where $\boldsymbol\beta = (\beta_0, \beta_1, \dots, \beta_{K+1})^\top$ is the vector of fixed-effects coefficients, $\mathbf b_i = (b_{0i}, b_{1i})^\top$ is the vector of random effects for subject $i$, and $\boldsymbol\varepsilon_i = (\varepsilon_{i1}, \dots, \varepsilon_{in_i})^\top$ is the vector of residual errors for subject $i$. 

The marginal distribution is
\[
  \mathbf y_i \;\sim\;
  \mathcal N\!\bigl(
      \mathbf X_i\boldsymbol\beta,\;
      \mathbf V_i\bigr),
  \qquad
  \mathbf V_i
  \;=\;
  \mathbf Z_i D \mathbf Z_i^\top + \sigma^2 I_{n_i}.
\]
Stacking all subjects gives
$\mathbf y\sim\mathcal N(\mathbf X\boldsymbol\beta,\mathbf V)$ with
$\mathbf V=\mathrm{blockdiag}(\mathbf V_1,\dots,\mathbf V_m)$.

The log‑likelihood can be written as:
\begin{equation}\label{eq:loglik}
  \ell(\boldsymbol\beta,\theta)
  \;=\;
  -\tfrac12\Bigl\{
     \log|\mathbf V|
     +(\mathbf y-\mathbf X\boldsymbol\beta)^\top\mathbf V^{-1}
       (\mathbf y-\mathbf X\boldsymbol\beta)
     + n\log(2\pi)
  \Bigr\},
\end{equation}
where $\theta=(\sigma_{b0}^2,\sigma_{b1}^2,\rho,\sigma^2)$ represents the vector of variance components. The estimation of $\boldsymbol\beta$ and $\theta$ via Maximum Likelihood (ML) proceeds as follows. Setting $\partial\ell/\partial\boldsymbol\beta=\mathbf0$ gives the generalised least‑squares (GLS) estimator for $\boldsymbol\beta$, conditional on $\theta$:
\begin{equation}\label{eq:beta_GLS}
  \hat{\boldsymbol\beta}
  \;=\;
  \bigl(\mathbf X^\top\mathbf V^{-1}\mathbf X\bigr)^{-1}
  \mathbf X^\top\mathbf V^{-1}\mathbf y.
\end{equation}
For each variance component $\theta_j$ in $\theta$, the ML estimate $\hat\theta_j$ is found by solving the score equation $\partial\ell/\partial\theta_j=\mathbf0$:
\[
  \frac{\partial\ell}{\partial\theta_j}
  \;=\;
  -\tfrac12
  \Bigl\{
     \operatorname{tr}\!\bigl(\mathbf V^{-1}\partial_{\theta_j}\mathbf V\bigr)
     -(\mathbf y-\mathbf X\boldsymbol\beta)^\top
        \mathbf V^{-1}(\partial_{\theta_j}\mathbf V)\mathbf V^{-1}
        (\mathbf y-\mathbf X\boldsymbol\beta)
  \Bigr\}=0.
\]
These equations are typically solved numerically (e.g., Newton–Raphson), often by iterating between estimating $\boldsymbol\beta$ given $\theta$, and then $\theta$ given $\boldsymbol\beta$, until convergence to obtain $\hat{\boldsymbol\beta}$ and $\hat\theta$.

While ML provides estimates for all parameters, its estimates of variance components ($\hat\theta$) can be biased, particularly in smaller samples, as ML does not fully account for the degrees of freedom used to estimate the fixed effects ($\boldsymbol\beta$).

Restricted Maximum Likelihood (REML) is preferred for estimating variance components as it yields less biased estimates. REML achieves this by maximizing a likelihood function based on linear combinations of $\mathbf y$ that are invariant to the fixed effects, effectively adjusting for the estimation of $\boldsymbol\beta$. The REML log-likelihood is:
\begin{equation}\label{eq:loglik_REML}
  \ell_{\text{REML}}(\theta)
  \;=\;
  -\tfrac12\Bigl\{
     \log|\mathbf V|
     +\log|\mathbf X^\top\mathbf V^{-1}\mathbf X|
     +(\mathbf y-\mathbf X\hat{\boldsymbol\beta})^\top
        \mathbf V^{-1}(\mathbf y-\mathbf X\hat{\boldsymbol\beta})
     +(n-p)\log(2\pi)
  \Bigr\},
\end{equation}
where $p=\dim(\boldsymbol\beta)$, and $\hat{\boldsymbol\beta}$ is the GLS estimator from \eqref{eq:beta_GLS}. REML estimates $\hat\theta_{\text{REML}}$ are found by solving $\partial\ell_{\text{REML}}/\partial\theta_j=\mathbf0$ numerically. Subsequently, $\boldsymbol\beta$ is estimated using GLS with $\mathbf V$ evaluated at $\hat\theta_{\text{REML}}$.
\subsection*{Numerical Estimation in Practice}\label{subsec:numerical_LMM}

Closed‑form solutions for the variance components $\theta$ do not exist,
so one resorts to iterative algorithms.  Two standard choices are
\emph{(i)} Newton\,/\,Fisher scoring on the log‑likelihood and
\emph{(ii)} the expectation–maximisation (EM) algorithm that treats the
random effects~$\mathbf b$ as latent variables,
which can
be implemented with standard software (e.g.\ \texttt{lme4} in \textsf{R}
or \texttt{statsmodels} in \textsc{Python}).

\subsubsection*{Variable Selection}
In modeling Parkinson’s disease progression, the presence of multicollinearity among voice biomarkers poses a significant challenge. Several acoustic features, such as the various jitter and shimmer measures, are known to be highly correlated (e.g., as shown in Figure \ref{fig:three_figs} (a), correlations exceeding 0.9 between some shimmer metrics). To address this and ensure model interpretability and parsimony, we employed a two-stage variable selection strategy: first using the Least Absolute Shrinkage and Selection Operator (LASSO) for dimensionality reduction, followed by backward stepwise selection on a linear mixed-effects model. Variance Inflation Factor (VIF) diagnostics were also computed to assess residual multicollinearity after selection.

The LASSO was applied on the linear model ignoring random effects, focusing solely on the fixed effects which shrinks less informative coefficients to zero, yielding a sparse set of candidate predictors. Importantly, the LASSO helped identify redundant jitter and shimmer variables, retaining only the most informative features for further modeling.

Subsequently, we performed stepwise backward selection using the \testtt{lmer} model, starting from a full linear mixed-effects model with all LASSO-selected predictors. This iterative process removed non-significant fixed effects based on AIC, leading to a reduced yet effective model. During this step, we also checked VIF values to confirm that no remaining variable exhibited severe multicollinearity (all VIF $< 5.0$). 

The final model retained five predictors: \testtt{age}, \testtt{test\_time}, \testtt{Jitter\_PPQ5}, \testtt{NHR}, and \testtt{HNR}. This subset balances interpretability, predictive performance, and model stability, and serves as the foundation for further modeling, including transformation, interaction terms, and random slopes. Note that interaction effects and nonlinear terms were considered in later modeling stages rather than during variable selection.

\subsubsection*{Model Refinement via Interaction and Random Slopes}

After identifying a subset of relevant predictors through variable selection, we further refined the linear mixed-effects model by incorporating interaction terms and evaluating the inclusion of subject-specific random slopes. These enhancements were designed to capture individual variation more flexibly and to model potential time-varying effects, thereby improving overall model fit and predictive performance.

\begin{figure}[ht]
    \centering
    \includegraphics[width=0.9\linewidth]{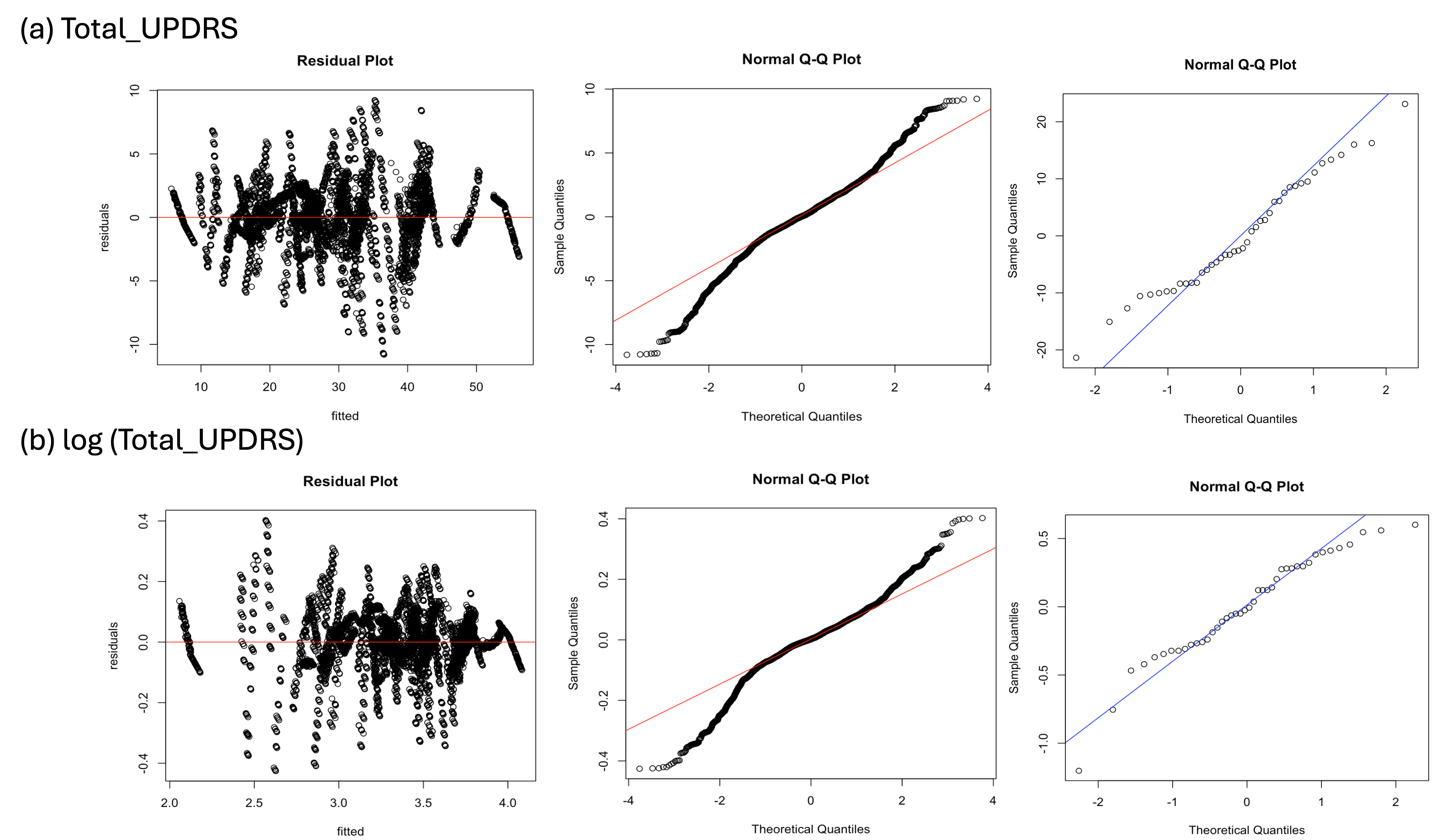}
    \caption{Top row: Residuals, fixed effect Q-Q, and random effect Q-Q plots from the original model using \texttt{total\_UPDRS}. Bottom row: Diagnostics after log-transforming the response. Transformation improves variance stabilization and normality.}
    \label{fig:transformation}
\end{figure}

To ensure the validity of the model assumptions, we examined diagnostic plots of residuals and normality. Figure \ref{fig:transformation} (top row) presents diagnostic plots from the initial model using the raw outcome variable \texttt{total\_UPDRS}. The residual plot reveals heteroscedasticity, with increasing spread at higher fitted values, while the Q-Q plots for both fixed and random effects show noticeable deviations from normality.

To mitigate these issues, we applied a logarithmic transformation to the outcome variable. This transformation significantly stabilized the variance and improved the normality of residuals, as shown in the bottom row of Figure \ref{fig:transformation}. It also compressed the scale of extreme values, reducing the influence of high-leverage tail points and producing more symmetric residuals overall.

In addition to transforming the response, we explored potential interactions among the selected covariates. A systematic evaluation of all pairwise interactions using likelihood-based model comparison revealed that the interaction between \texttt{test\_time} and \texttt{HNR} was the most impactful. This interaction was statistically significant and led to a notable improvement in model fit, with the AIC decreasing from $-10231.6$ to $-10243.3$. The result suggests that the effect of \texttt{HNR} on disease progression varies over time. Other interactions provided marginal improvement or introduced unnecessary complexity and were therefore excluded from the final model.

We also investigated whether to include random slopes in addition to random intercepts for each subject. As shown in Figure \ref{fig:three_figs} (c), subject-specific log(UPDRS) trajectories over time exhibited heterogeneous slopes, motivating the inclusion of a random slope for \testtt{test\_time}. The addition of this random slope further reduced AIC to $-10261.4$, resulting in the final model:
\begin{align*}
\log(\text{UPDRS}*{ij}) &= \beta_0 + \beta_1 \text{age}_i + \beta_2 \text{test\_time}_{ij} + \beta_3 \text{HNR}_{ij} + \beta_4 (\text{test\_time}_{ij} \times \text{HNR}_{ij}) \\
&+ b_{0i} + b_{1i} \text{test\_time}_{ij} + \varepsilon_{ij}
\end{align*}
where $b_{0i}, b_{1i} \sim \mathcal{N}(0, G)$ represent the subject-specific random intercept and slope for time, and $\varepsilon_{ij} \sim \mathcal{N}(0, \sigma^2)$ denotes the residual error.

Table \ref{tab:lmm1} summarizes the model refinement process using AIC as the selection criterion. Each modification, i.e., the selection of variables, the transformation, the inclusion of interactions and the random slope, improved the model fit, culminating in the final model specification above.

\begin{table}[ht]
\centering
\small
\caption{\footnotesize Model selection and refinement steps based on AIC comparison.}
\begin{tabular}{lcc}
\toprule
\textbf{Step} & \textbf{AIC} & \textbf{Comment} \\
\midrule
Full model (all predictors)   & 28124.1   & Initial LMM \\
After LASSO (fixed effects only) & 27880.6        & Removed highly correlated terms \\
After stepwise (LMM) \& VIF & 27869.8   & Dropped non-significant effects \\
After log-transformation      & –10231.6  & Improved residual normality, variance \\
Add interaction (test\_time $\times$ HNR)       & –10243.3  & Included significant time-varying HNR effect \\
Add random slope              & \textbf{–10261.4} & Final model with varying subject-specific slopes \\
\bottomrule
\end{tabular}
\label{tab:lmm1}
\end{table}

This finalized model serves as a baseline for comparison against more flexible methods, such as generalized additive and deep mixed-effects models, in subsequent sections.


\subsection{Generalized Additive Mixed Model (GAMM)}\label{sec:gamm}

To capture nonlinear temporal effects in Parkinson’s disease progression, we adopted a Generalized Additive Mixed Model (GAMM), which extends the linear mixed-effects framework by allowing smooth, data-driven functions of continuous covariates \citep{wood2017generalized}. This formulation maintains the interpretability of linear effects while introducing the flexibility necessary to model nonlinear trends over time.

In our application, the log-transformed total UPDRS score is modeled as a smooth function of \texttt{test\_time}, along with linear terms for other covariates. The model is expressed as:
\begin{align}
  \log(\text{UPDRS}_{ij}) &= \beta_0 + \beta_1 \text{age}_i + \beta_2 \text{HNR}_{ij} + f(\texttt{test\_time}_{ij}) + b_{0i} + b_{1i} \texttt{test\_time}_{ij} + \varepsilon_{ij}, \label{eq:gamm}
\end{align}
where $f(\cdot)$ is a smooth function of time, and $b_{0i}, b_{1i} \sim \mathcal{N}(0, G)$ are the subject-specific random intercept and slope. The residual errors $\varepsilon_{ij} \sim \mathcal{N}(0, \sigma^2)$ are assumed to be independent and homoscedastic.

\subsubsection*{Spline Basis Representation and Estimation}

The smooth function $f(\texttt{test\_time})$ is approximated via a linear combination of basis functions:
\begin{align}
f(\texttt{test\_time}) = \sum_{k=1}^K \alpha_k B_k(\texttt{test\_time}), \label{eq:spline_basis}
\end{align}
where $B_k(\cdot)$ are predefined spline basis functions (e.g., cubic regression splines, B-splines, or thin plate splines \citep{ruppert2003semiparametric}), and $\alpha_k$ are coefficients estimated from the data. To control smoothness and avoid overfitting, a roughness penalty is imposed on the second derivative of the function:
\begin{align}
\text{Penalized log-likelihood} = \ell(\boldsymbol{\beta}, \boldsymbol{\alpha}) - \frac{1}{2} \lambda \int \left[f''(t)\right]^2 dt, \label{eq:spline_penalty}
\end{align}
where $\lambda$ is a smoothing parameter that balances model fit and regularity. The complexity of the spline is quantified via its effective degrees of freedom (edf).

Estimation proceeds using Penalized Iteratively Reweighted Least Squares (P-IRLS), an efficient approach for maximizing the penalized likelihood. The \texttt{gamm()} function in the \texttt{mgcv} R package is used to jointly estimate the fixed effects, the smooth term, and the random effects. Internally, \texttt{gamm()} delegates the random effects estimation to the \texttt{lme()} function from the \texttt{nlme} package, allowing for flexible modeling of subject-specific deviations via random intercepts and slopes.

Smoothing parameters \( \lambda \) are selected by optimizing the marginal Restricted Maximum Likelihood (REML) criterion, which balances model fit with smoothness and has been shown to offer stability and efficiency in practice \citep{wood2011fast}.

\subsubsection*{Comparison of LMM and GAMM}

To assess model performance, we compared the final Linear Mixed-Effects Model (LMM) and the Generalized Additive Mixed Model (GAMM) using both model fit criteria and predictive accuracy. Table~\ref{tab:model_comparison} summarizes the estimated fixed and random effects, along with model fit and test set performance.

\begin{figure}[ht]
\centering
\includegraphics[width=0.6\linewidth]{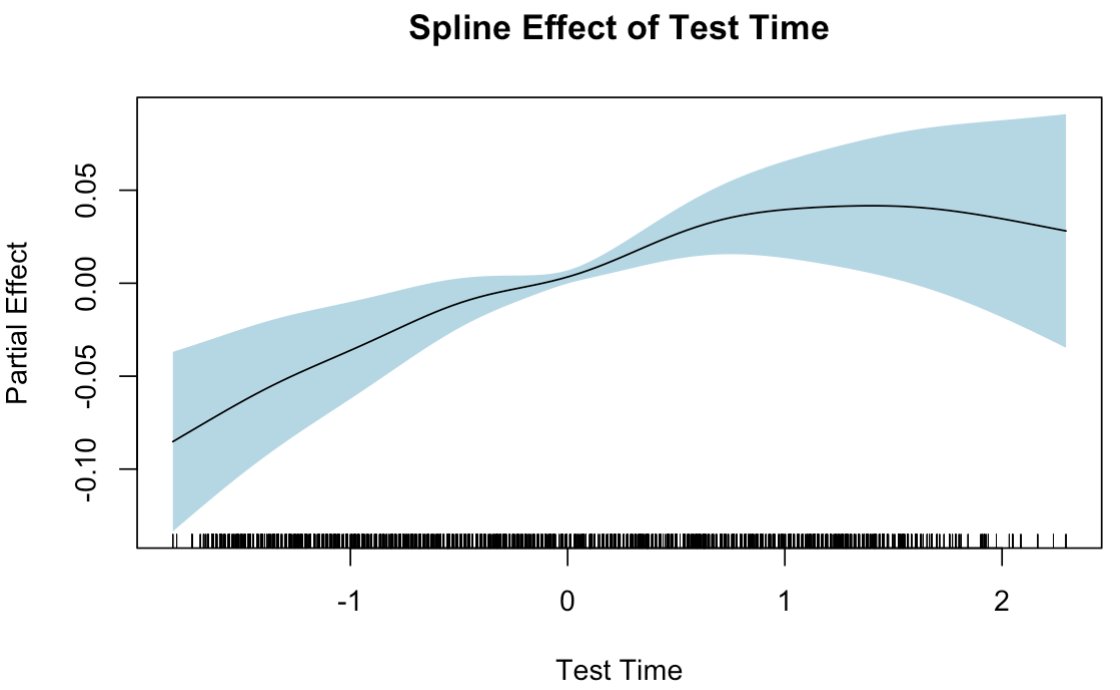}
\caption{Estimated spline effect of \texttt{test\_time} from GAMM, showing nonlinear progression over time.}
\label{fig:spline_plot}
\end{figure}

The GAMM demonstrated a superior model fit based on Akaike Information Criterion (AIC), achieving a lower AIC value ($-16162.05$) than the LMM ($-16062.39$). This improvement can be attributed to GAMM's flexibility in capturing nonlinear structures, particularly through a spline term applied to \texttt{test\_time}. As shown in Figure~\ref{fig:spline_plot}, the estimated smooth function of time deviates notably from linearity, reinforcing the presence of nonlinear progression patterns in UPDRS scores over time.

To compare predictive performance, we conducted a hold-out evaluation: the last observation from each of the 42 subjects was set aside as the test set, while the remaining data were used for training. On this test set, GAMM achieved a slightly lower mean squared error (MSE = 6.56) compared to LMM (MSE = 7.70), indicating marginally better prediction accuracy in the original scale.

Notably, the interaction term \texttt{test\_time $\times$ HNR} was statistically significant in the LMM but became non-significant under GAMM. This suggests that the nonlinear main effect of \texttt{test\_time} in GAMM may account for variation previously explained by the interaction term in the LMM. The estimated standard deviations of the random effects and residual terms were similar across models, indicating consistent subject-level variation.

\begin{table}[ht]
\centering
\caption{Comparison of LMM and GAMM Estimates and Performance. The AIC value for LMM reported here differs from Table \ref{tab:lmm1} because the final models in this comparison were fitted using Restricted Maximum Likelihood (REML) for unbiased variance estimation, whereas the model selection process in Table \ref{tab:lmm1} utilized Maximum Likelihood (ML) to compare fixed effects.}
\label{tab:model_comparison}
\small
\begin{tabular}{lccc}
\toprule
\textbf{Fixed Effects (Est. (p-val))} & \textbf{LMM} & \textbf{GAMM} & Notes\\
\midrule
Intercept & 3.316 ($<$2e-16) & 3.316 ($<$2e-16) & \\
Age       & 0.136 (0.0219) & 0.136 (0.0145) & Significant in both \\
test\_time & 0.0358 (0.0099) & spline & Nonlinear in GAMM \\
HNR       & 0.0036 (0.0035) & 0.0043 (0.0006) & Consistent positive effect\\
test\_time:HNR & -0.0077 (1e-08) & n.s. & Only significant in LMM \\
\midrule
\textbf{Smooth Terms (GAMM only)} & \multicolumn{3}{c}{s(test\_time): edf = 6.17, p $<$ 2e-16} \\
\midrule
\textbf{Random Effects (Std. Dev (Corr))} & & & \\
Intercept   & 0.382 & 0.373 & Similar \\
Slope (test\_time) & 0.085 ($-0.08$) & 0.084 ($-0.061$) & Similar \\
Residual    & 0.058 & 0.057 & Similar \\
\midrule
\textbf{Model Fit (AIC)} & -16062.39 & \textbf{-16162.05} & GAMM better (fit) \\
\midrule
\textbf{Test MSE (orig. scale)} & 7.70 & \textbf{6.56} & GAMM better (predictive) \\
\bottomrule
\end{tabular}
\end{table}

In summary, GAMM provided a more flexible fit by capturing nonlinear temporal patterns, as evident in its superior AIC and spline visualization. However, LMM offered comparable generalization performance on the test set and more interpretable fixed effects. The choice between models thus depends on whether the goal prioritizes interpretability or modeling flexibility.


\subsection{Generalized Neural Network Mixed Model (GNMM) for Non‑linear Longitudinal Modeling}

\vspace{0.3em}
Building on Mandel \cite{Mandel2021} we use a Generalized Neural Network Mixed Model (GNMM) to predict the longitudinal \texttt{Total\_UPDRS} scores collected in the tele‑monitoring study of Parkinson's disease.  

We retain the notation introduced earlier: $Y_{ij}$ is the \texttt{Total\_UPDRS} score for subject $i$ at visit $j$, $\mathbf X_{ij}\in\mathbb R^{p}$ is the $p=17$‑vector of predictors (test time + 16 voice features).

Let \(i=1,\dots,m\) label the \(m=42\) patients in the Oxford telemonitoring study and \(j=1,\dots,n_i\) their successive visits.  At visit \(j\) we record the response \(Y_{ij}\) (the \texttt{Total\_UPDRS} score) and a predictor vector
\[
\mathbf X_{ij}=(\texttt{test\_time},\text{16 voice features})^{\!\top}\in\mathbb R^{17},
\]
where \(\texttt{test\_time}\) is the elapsed study time and the remaining 16 entries are acoustic measures extracted from the voice recording.

Following the mixed‑effects formulation of \cite{Mandel2021} , we allow observations from the same subject to be correlated through a \emph{cluster‑specific random‑effect vector} $\mathbf b_i\in\mathbb R^{q}$, where $q\ge 1$. 
Conditional on $\mathbf b_i$, the outcomes $Y_{ij}$ are assumed to follow an exponential‑family distribution
\[
E\!\bigl[Y_{ij}\mid\mathbf b_i\bigr]=\mu_{ij}^{\mathbf b},
\qquad
\operatorname{Var}\!\bigl(Y_{ij}\mid\mathbf b_i\bigr)=\phi\,a_{ij}\,v\!\bigl(\mu_{ij}^{\mathbf b}\bigr),
\]
with known variance function $v(\cdot)$, fixed dispersion $\phi$, and  $a_{ij}$ is a known constant.

\subsection*{Generalized Neural Network Mixed Model (GNMM) On Our Case}

Consider a feed‑forward artificial neural network (ANN) with $L$ hidden layers, the predictor vector $\mathbf X_{ij}\in\mathbb R^{p}$ ($p=17$) as input, and a univariate output $\mu^{\mathbf b}_{ij}$ representing the conditional mean of $Y_{ij}$ (\texttt{Total\_UPDRS}) for subject $i$ at visit $j$.  Following \cite{Mandel2021}, the network output is built up through a sequence of nested activation functions $g_{\ell}(\cdot)$, $\ell=0,\dots,L$.

\subsection*{Network layers.}
The input $\mathbf X_{ij}$ enters the $L$‑th (bottom) hidden layer with $k_{L}$ nodes:
\begin{equation}
\boldsymbol\alpha^{(L)}_{ij}=g_{L}\!\bigl\{\boldsymbol\omega^{(L)}\mathbf X_{ij}+\boldsymbol\delta^{(L)}\bigr\},
\label{eq:GNMM_layerL}
\end{equation}
where $\boldsymbol\omega^{(L)}$ is a $k_{L}\times p$ weight matrix and $\boldsymbol\delta^{(L)}$ is a bias vector of length $k_{L}$.  For hidden layer $\ell=L-1,\dots,1$ with $k_{\ell}$ nodes,
\begin{equation}
\boldsymbol\alpha^{(\ell)}_{ij}=g_{\ell}\!\bigl\{\boldsymbol\omega^{(\ell)}\boldsymbol\alpha^{(\ell+1)}_{ij}+\boldsymbol\delta^{(\ell)}\bigr\},
\label{eq:GNMM_layer}
\end{equation}
with $\boldsymbol\omega^{(\ell)}$ of size $k_{\ell}\times k_{\ell+1}$ and $\boldsymbol\delta^{(\ell)}\in\mathbb R^{k_{\ell}}$.

\subsection*{Output layer and random effects.}
The univariate network output determines the conditional mean through
\begin{equation}
\mu^{\mathbf b}_{ij}=g_{0}\!\bigl\{\boldsymbol\omega^{(0)}\boldsymbol\alpha^{(1)}_{ij}+\delta^{(0)}+\mathbf Z_{ij}^{\top}\mathbf b_{i}\bigr\},
\label{eq:GNMM_mean}
\end{equation}
where $\boldsymbol\omega^{(0)}$ is a $1\times k_{1}$ weight vector, $\delta^{(0)}$ a scalar bias, $\mathbf Z_{ij}\in\mathbb R^{q}$ the design vector for the cluster‑specific random effect, and $\mathbf b_{i}\sim N(\mathbf 0,\mathbf D)$.  In a classical generalized linear mixed model \cite{Breslow1993},
\begin{equation}
E\!\bigl[Y_{ij}\mid\mathbf b_i\bigr]
  = h\!\bigl(\mathbf X_{ij}^{\top}\boldsymbol\alpha
             + \mathbf Z_{ij}^{\top}\mathbf b_i\bigr),
\label{eq:GLMM_standard}
\end{equation}
where $h(\cdot)$ is the inverse link, $\mathbf X_{ij}^{\top}\boldsymbol\alpha$ the fixed‑effect component, and $\mathbf Z_{ij}^{\top}\mathbf b_i$ the random effect.  
In the GNMM we replace the fixed‑effect term by the nonlinear network output and use the final activation $g_{0}(\cdot)$ in place of $h(\cdot)$:
\begin{equation}
E\!\bigl[Y_{ij}\mid\mathbf b_i\bigr]
  = g_{0}\!\bigl(
      \boldsymbol\omega^{(0)}\boldsymbol\alpha^{(1)}_{ij}
      + \delta^{(0)}
      + \mathbf Z_{ij}^{\top}\mathbf b_i
      \bigr).
\label{eq:GNMM_replacement}
\end{equation}

\subsection*{Quasi‑likelihood.}
Let $\boldsymbol\omega=\operatorname{vec}\!\bigl(\boldsymbol\omega^{(0)},\boldsymbol\omega^{(1)}\bigr)$
and $\boldsymbol\delta=\bigl(\delta^{(0)},\boldsymbol\delta^{(1)}\bigr)$
collect all weights and biases of the single‑hidden–layer network.
With $q=1$ (a random intercept) we set
$\mathbf Z_{ij}\equiv1$ and $b_i\sim N(0,D)$.
The quasi‑likelihood used to estimate
$\bigl(\boldsymbol\omega,\boldsymbol\delta,D\bigr)$ is
\begin{equation}
\exp\!\{ql(\boldsymbol\omega,\boldsymbol\delta,\theta)\}\;\propto\;
|D|^{-1/2}
\int
\exp\Biggl\{
  \frac{1}{\phi}\sum_{i=1}^{42}\sum_{j=1}^{n_i}
     \int_{Y_{ij}}^{\mu^{b}_{ij}}
       \frac{y_{ij}-u}{a_{ij}v(u)}\,\mathrm du
  \;-\;
  \frac{b^{2}}{2D}
  \;-\;
  \lambda\!\bigl(\boldsymbol\omega^{\!\top}\boldsymbol\omega
                +\boldsymbol\delta^{\!\top}\boldsymbol\delta\bigr)
\Biggr\}\,
\mathrm db .
\label{eq:ql_integral}
\end{equation}

\subsection*{Laplace approximation.}
Following \cite{Breslow1993} and \cite{Mandel2021}, write
\[
\kappa(b)=
  -\sum_{j=1}^{n_i}\!\int_{Y_{ij}}^{\mu^{b}_{ij}}
      \frac{y_{ij}-u}{\phi\,a_{ij}v(u)}\,\mathrm du
  + \frac{b^{2}}{2D}
  + \frac{\lambda}{2}\bigl(\boldsymbol\omega^{\!\top}\boldsymbol\omega
                           +\boldsymbol\delta^{\!\top}\boldsymbol\delta\bigr),
\]
so that \eqref{eq:ql_integral} is proportional to
$|D|^{-1/2}\!\int\!\exp\!\{-\kappa(b)\}\,\mathrm db$.
Let $\tilde b$ be the mode obtained from
$\partial\kappa/\partial b=0$; first‑ and second‑order derivatives give
\[
\kappa'(b)=
  -\sum_{j=1}^{n_i}
     \frac{(Y_{ij}-\mu^{b}_{ij})
           g_{0}'\!\bigl(\eta_{ij}^{b}\bigr)}%
          {\phi\,a_{ij}v\!\bigl(\mu^{b}_{ij}\bigr)}
  + \frac{b}{D},
\qquad
\kappa''(b)=
  Z_{i}^{\top}W_{i}Z_{i}+D^{-1},
\]
where $Z_{i}=\mathbf 1_{n_i}$ and
$W_{i}=\operatorname{diag}\!\bigl\{\phi^{-1}
        a_{ij}^{-1}v\!\bigl(\mu^{\tilde b}_{ij}\bigr)^{-1}
        g_{0}'\!\bigl(\eta_{ij}^{\tilde b}\bigr)^{2}\bigr\}$.
Ignoring the remainder term yields the Laplace approximation of the Quasi-likelihood function, which will be used in the following steps.
\begin{equation}
ql(\boldsymbol\omega,\boldsymbol\delta,\theta)
\;\approx\;
-\frac12\log|D|
-\frac12\log\bigl(Z_{i}^{\top}W_{i}Z_{i}+D^{-1}\bigr)
-\frac{\lambda}{2}\bigl(\boldsymbol\omega^{\!\top}\boldsymbol\omega
                        +\boldsymbol\delta^{\!\top}\boldsymbol\delta\bigr).
\label{eq:ql_laplace}
\end{equation}

\subsection*{Training objective.}
Treating the $W_{i}$ term in \eqref{eq:ql_laplace} as negligible
(as in \cite{Mandel2021}) leads to the objective maximised during
network training:
\begin{equation}
ql(\boldsymbol\omega,\boldsymbol\delta,\theta)
\;\propto\;
\frac{1}{\phi}
\sum_{i=1}^{42}\sum_{j=1}^{n_i}
  \int_{Y_{ij}}^{\mu^{\tilde b}_{ij}}
    \frac{y_{ij}-u}{a_{ij}v(u)}\,\mathrm du
\;-\;
\frac{\tilde b^{2}}{2D}
\;-\;
\lambda\!\bigl(\boldsymbol\omega^{\!\top}\boldsymbol\omega
              +\boldsymbol\delta^{\!\top}\boldsymbol\delta\bigr).
\label{eq:ql_training}
\end{equation}

\subsection*{Quasi‑score equations (single hidden layer).}
Differentiating \eqref{eq:ql_training} with respect to the network
parameters yields the quasi‑score system
\begin{align}
\frac{\partial ql}{\partial \omega^{(0)}_{k}}
 &= \frac{1}{\phi}\sum_{i=1}^{42}\sum_{j=1}^{n_i}
      (Y_{ij}-\mu^{\tilde b}_{ij})\,
      g_{0}'\!\bigl(\eta_{ij}^{\tilde b}\bigr)\,
      \alpha^{(1)}_{ij,k}
    - 2\lambda\,\omega^{(0)}_{k}, \tag{QS1}\\[0.4em]
\frac{\partial ql}{\partial \delta^{(0)}}
 &= \frac{1}{\phi}\sum_{i=1}^{42}\sum_{j=1}^{n_i}
      (Y_{ij}-\mu^{\tilde b}_{ij})\,
      g_{0}'\!\bigl(\eta_{ij}^{\tilde b}\bigr)
    - 2\lambda\,\delta^{(0)}, \tag{QS2}\\[0.4em]
\frac{\partial ql}{\partial \omega^{(1)}_{lk}}
 &= \frac{1}{\phi}\sum_{i=1}^{42}\sum_{j=1}^{n_i}
      (Y_{ij}-\mu^{\tilde b}_{ij})\,
      g_{0}'\!\bigl(\eta_{ij}^{\tilde b}\bigr)\,
      \omega^{(0)}_{k}\,
      g_{1}'\!\bigl(s_{ij,k}\bigr)\,X_{ij,l}
    - 2\lambda\,\omega^{(1)}_{lk}, \tag{QS3}\\[0.4em]
\frac{\partial ql}{\partial \delta^{(1)}_{k}}
 &= \frac{1}{\phi}\sum_{i=1}^{42}\sum_{j=1}^{n_i}
      (Y_{ij}-\mu^{\tilde b}_{ij})\,
      g_{0}'\!\bigl(\eta_{ij}^{\tilde b}\bigr)\,
      \omega^{(0)}_{k}\,
      g_{1}'\!\bigl(s_{ij,k}\bigr)
    - 2\lambda\,\delta^{(1)}_{k}. \tag{QS4}
\end{align}

\noindent
Here $\eta_{ij}^{\tilde b}
 = \boldsymbol\omega^{(0)}\boldsymbol\alpha^{(1)}_{ij}
   +\delta^{(0)}+\tilde b_i$,
$\displaystyle
s_{ij,k}=(\boldsymbol\omega^{(1)}_{k})^{\!\top}\mathbf X_{ij}
          +\delta^{(1)}_{k}$,
and $\alpha^{(1)}_{ij,k}=g_{1}(s_{ij,k})$.
Equation \eqref{eq:ql_training} is maximised by solving
(QS1)–(QS4) jointly with $\kappa'(\tilde b)=0$;
we use back‑propagation for the network parameters and
stochastic gradient descent on $\tilde b$,
updating $\tilde b$ at each epoch while keeping $D$
fixed at its REML estimate (as in GLMM, with
$\mathbf X_{ij}^{\top}\boldsymbol\beta$ replaced by
$\boldsymbol\omega^{(0)}\boldsymbol\alpha^{(1)}_{ij}+\delta^{(0)}$).

\subsection*{Summary of the Algorithm}

The GNMM network transforms the 17‑dimensional vector of voice features into
a latent disease score, allowing nonlinear interactions and saturation
effects to influence the predicted \texttt{Total\_UPDRS}.  The random
intercept $b_{i}$ absorbs persistent patient‑level deviations, so
estimates borrow strength across subjects (partial pooling).  An
$L_{2}$ penalty
$\lambda(\boldsymbol\omega^{\top}\boldsymbol\omega
        +\boldsymbol\delta^{\top}\boldsymbol\delta)$
decreases large weights and biases, mitigating over‑fitting.

The algorithm flow used is summarzied as below:
\begin{algorithm}[H]
\caption{Stochastic training of GNMM on the Parkinson data (adapted from \cite{Mandel2021})}
\label{alg:gnmm_training}
\begin{algorithmic}[1]
\Require scaled features $\mathbf X'$, scaled targets $Y'$, subject indices; epochs $E$, batch size $B$, learning rate $\eta$
\State Initialise parameters $\boldsymbol\vartheta$ (Xavier), $b_i\gets0$, $\sigma^{2}\gets1$, $\sigma_{b}^{2}\gets1$
\For{$e=1,\dots,E$}
    \State Shuffle the training set
    \For{each mini‑batch $\mathcal B$ of size $B$}
        \State Compute $\mu_{ij}^{\,b}$ for $(i,j)\in\mathcal B$ via (\ref{eq:GNMM_mean})
        \State Evaluate mini‑batch loss $L_{\mathcal B}$ from (\eqref{eq:ql_training}
        \State Back‑propagate $\nabla_{\boldsymbol\vartheta}L_{\mathcal B}$ and $\nabla_{b}L_{\mathcal B}$
        \State Update $\boldsymbol\vartheta \leftarrow \boldsymbol\vartheta - \eta\,\nabla_{\boldsymbol\vartheta}L_{\mathcal B}$
        \State Update each $b_i$ that appears in $\mathcal B$
    \EndFor
    \State $\sigma^{2}\leftarrow$ mean squared residual over the full training set
    \State $\sigma_{b}^{2}\leftarrow$ sample variance of $\{b_i\}$
\EndFor
\end{algorithmic}
\end{algorithm}

\subsection*{Implementation and Results}

We implemented the GNMM in \textsf{R} using the \texttt{gnmm.sgd} and \texttt{gnmm.predict} routines provided in the Supplementary Material of \cite{Mandel2021} and tailored for our own dataset. Following the evaluation strategy defined for the LMM and GAMM, the final visit for each of the 42 subjects was held out as the test set.

We compare the following models:
\begin{itemize}
    \item \emph{1‑layer GNMM:} one hidden layer with three \texttt{ReLU} nodes,
          ridge penalty $\lambda=0.001$, learning rate $0.005$,
          random intercept included.
    \item \emph{2‑layer GNMM:} two hidden layers (three and two nodes),
          $\lambda=0.002$, learning rate $0.005$, random intercept included.
    \item \emph{ANN baseline:} one hidden layer with three nodes,
          $\lambda=0.001$, learning rate $0.001$, \emph{no} random effect.
\end{itemize}

\paragraph{Evaluation Strategy and Limitations.}
Predictive accuracy was measured on the held-out visits using mean squared error (MSE) and mean absolute error (MAE). To account for initialization variability in high-capacity models, each model was trained \textbf{five times} with independent random seeds; Table \ref{tab:gnmm_results} reports the average performance.

Following the strategy of \cite{Mandel2021}, the final visit for every subject was held out for testing. It is important to clarify that this split evaluates the model's ability to \textit{forecast} disease trajectory for patients already under monitoring (interpolation), rather than \textit{generalizing} to entirely new subjects (extrapolation). We also acknowledge that the data magnitude presents a specific challenge for stability. While the total number of observations is substantial ($N_{obs} = 5,875$), the effective sample size for estimating subject-specific random effects is constrained to the number of patients ($N=42$). This creates a high parameter-to-sample ratio for the neural mixed-effects models, particularly the NME which optimizes non-linear parameters across a multi-layer architecture. To rigorously assess robustness in this small-$N$ regime, we report results averaged over the five independent training runs to mitigate initialization bias.

\begin{table}[H]
\centering
\caption{Average test‑set error over five independent runs}
\begin{tabular}{lcc}
\toprule
Model & MSE & MAE \\
\midrule
1‑layer GNMM & $96.82$ & $6.96$ \\
2‑layer GNMM & $106.09$ & $7.56$ \\
ANN (no random effect) & $114.20$ & $8.47$ \\
\bottomrule
\end{tabular}
\label{tab:gnmm_results}
\end{table}

\noindent
The single‑layer GNMM attains the lowest prediction error, reducing
test‑set MSE by $15\%$ relative to the two‑layer variant and by
$15.3\%$ relative to the feed‑forward network without random effects.

\subsection{Neural Mixed-Effects (NME) Model for Longitudinal UPDRS Prediction}

Another recently introduced neural network model that can be applied to our case is the Neural Mixed-Effects (NME) model.

Classical mixed-effects models (LMM, GLMM) effectively handle subject heterogeneity in longitudinal data but are typically restricted to linear fixed effects. Conversely, standard neural networks can learn rich nonlinear relationships but often ignore the within-subject correlation inherent in repeated measures. The Neural Mixed Effects (NME) framework, as proposed by Wörtwein \emph{et al.} \cite{Wortwein2023}, elegantly combines these strengths. This framework permits the inclusion of nonlinear subject-specific parameters at any layer of the network and utilizes stochastic gradient descent for optimization, which ensures scalability with both the number of patients ($m$) and the total number of visits (i.e., $\sum_{i=1}^{m} n_i$).

Applying the NME approach to our Parkinson's tele-monitoring study offers several  advantages. First of all, it allows for the learning of complex relationships between voice features and disease severity without the need for pre-specifying  interaction terms.

Additionally, the NME model employs partial pooling for its parameter estimates. This approach allows the model to share information across different patients, leading to more robust and reliable individual-specific parameters, particularly for patients with fewer observations, by balancing individual data with overall population trends.

\subsection*{NME Parameterization.}
Let $i=1,\dots,m$ index the $m=42$ participants in the Parkinson's tele-monitoring study, and $j=1,\dots,n_i$ index their repeated visits. At each visit $j$ for participant $i$, we observe the response $Y_{ij}$, representing the UPDRS score, and a $p$-dimensional predictor vector $\mathbf{X}_{ij} \in \mathbb{R}^{17}$ which is  consisted of test time (time of assessment) and 16 scaled voice features.
Following Wörtwein \emph{et al.} \cite{Wortwein2023}, the NME model decomposes the network parameters into two components:
\begin{enumerate}
    \item A \emph{person-generic} component $\bar{\boldsymbol{\theta}}$, which is shared across all participants and captures common trends.
    \item A \emph{person-specific} component $\boldsymbol{\eta}_i$, unique to participant $i$, capturing individual deviations from the generic trend.
\end{enumerate}
The effective parameters for participant $i$ are thus $\boldsymbol{\theta}_i = \bar{\boldsymbol{\theta}} + \boldsymbol{\eta}_i$. The person-specific components $\boldsymbol{\eta}_i$ are typically regularized by assuming they follow a multivariate normal distribution, $\boldsymbol{\eta}_i \sim N(\mathbf{0}, \boldsymbol{\Sigma})$, where $\boldsymbol{\Sigma}$ is a covariance matrix (often diagonal, e.g., $\boldsymbol{\Sigma} = \tau^2\mathbf{I}$).

\subsection*{Network Architecture for UPDRS Prediction.}
To capture the complex mapping between voice features and disease severity, we employ a multilayer perceptron (MLP) architecture. Unlike standard MLPs, our implementation integrates the mixed-effects structure directly into the network weights. This allows the model to learn a global population trend (via person-generic parameters) while simultaneously adjusting weights and biases for each individual (via person-specific deviations), thereby personalizing the prediction for each patient.

Specifically, we implement a \textbf{two-hidden-layer MLP} with $k_1=32$ units in the first hidden layer and $k_2=16$ units in the second hidden layer. The parameters of this network are decomposed into person-generic components (elements of $\bar{\boldsymbol{\theta}}$) and person-specific deviations (elements of $\boldsymbol{\eta}_i$) as defined in Equation~\eqref{eq:NME_param_split_combined}.
For predicting \texttt{total\_UPDRS}, we implement a \textbf{two-hidden-layer multilayer perceptron (MLP)} with $k_1=32$ units in the first hidden layer and $k_2=16$ units in the second hidden layer. The parameters of this network are decomposed into person-generic components (elements of $\bar{\boldsymbol{\theta}}$) and person-specific deviations (elements of $\boldsymbol{\eta}_i$) as defined in Equation~\eqref{eq:NME_param_split_combined}. Specifically, the network operations involving these decomposed parameters are defined as:
\begin{align}
\boldsymbol{\alpha}^{(1)}_{ij} &= g_{1a}\!\left((\bar{\boldsymbol{\Omega}}^{(1)} + \boldsymbol{\eta}_{\Omega^{(1)},i}) \mathbf{X}_{ij} + (\bar{\boldsymbol{\delta}}^{(1)} + \boldsymbol{\eta}_{\delta^{(1)},i})\right) \label{eq:NME_hidden1_activation_combined_final} \\
\boldsymbol{\alpha}^{(2)}_{ij} &= g_{1b}\!\left((\bar{\boldsymbol{\Omega}}^{(2)} + \boldsymbol{\eta}_{\Omega^{(2)},i}) \boldsymbol{\alpha}^{(1)}_{ij} + (\bar{\boldsymbol{\delta}}^{(2)} + \boldsymbol{\eta}_{\delta^{(2)},i})\right) \label{eq:NME_hidden2_activation_combined_final} \\
\hat{Y}_{ij} \equiv \mu_{ij}^{\text{NME}} &= g_{0}\!\left((\bar{\boldsymbol{\omega}}^{(0)} + \boldsymbol{\eta}_{\omega^{(0)},i}) \boldsymbol{\alpha}^{(2)}_{ij} + (\bar{\delta}^{(0)} + \boldsymbol{\eta}_{\delta^{(0)},i})\right) \label{eq:NME_output_two_hidden_combined_final}
\end{align}
In these equations, $\mathbf{X}_{ij}$ represents the input features for subject $i$ at visit $j$. The term $\boldsymbol{\alpha}^{(1)}_{ij}$ denotes the activations of the first hidden layer. These are computed using the input-to-first-hidden-layer person-generic weights $\bar{\boldsymbol{\Omega}}^{(1)}$ (a component of $\bar{\boldsymbol{\theta}}$) and person-specific weight deviations $\boldsymbol{\eta}_{\Omega^{(1)},i}$ (a component of $\boldsymbol{\eta}_i$), as well as the corresponding biases $\bar{\boldsymbol{\delta}}^{(1)}$ and $\boldsymbol{\eta}_{\delta^{(1)},i}$, followed by the activation function $g_{1a}(\cdot)$.
Similarly, $\boldsymbol{\alpha}^{(2)}_{ij}$ represents the activations of the second hidden layer, taking $\boldsymbol{\alpha}^{(1)}_{ij}$ as input. This layer uses person-generic weights $\bar{\boldsymbol{\Omega}}^{(2)}$ and biases $\bar{\boldsymbol{\delta}}^{(2)}$, with their respective person-specific deviations $\boldsymbol{\eta}_{\Omega^{(2)},i}$ and $\boldsymbol{\eta}_{\delta^{(2)},i}$, followed by its activation function $g_{1b}(\cdot)$. For both hidden layers, the activation function used is the Rectified Linear Unit (ReLU).
The final prediction, $\hat{Y}_{ij}$ (or $\mu_{ij}^{\text{NME}}$), is obtained from the output layer. This layer takes $\boldsymbol{\alpha}^{(2)}_{ij}$ as input and applies the second-hidden-to-output-layer person-generic weights $\bar{\boldsymbol{\omega}}^{(0)}$ and output biases $\bar{\delta}^{(0)}$, along with their person-specific deviations $\boldsymbol{\eta}_{\omega^{(0)},i}$ and $\boldsymbol{\eta}_{\delta^{(0)},i}$. The output layer activation function $g_0(\cdot)$ is the identity function ($g_0(x)=x$), as \texttt{total\_UPDRS} is a continuous response.
Collectively, the person-generic parameters are $\bar{\boldsymbol{\theta}} = (\bar{\boldsymbol{\Omega}}^{(1)}, \bar{\boldsymbol{\delta}}^{(1)}, \bar{\boldsymbol{\Omega}}^{(2)}, \bar{\boldsymbol{\delta}}^{(2)}, \bar{\boldsymbol{\omega}}^{(0)}, \bar{\delta}^{(0)})$, and the person-specific deviations for subject $i$ are $\boldsymbol{\eta}_i = (\boldsymbol{\eta}_{\Omega^{(1)},i}, \boldsymbol{\eta}_{\delta^{(1)},i}, \boldsymbol{\eta}_{\Omega^{(2)},i}, \boldsymbol{\eta}_{\delta^{(2)},i}, \boldsymbol{\eta}_{\omega^{(0)},i}, \boldsymbol{\eta}_{\delta^{(0)},i})$. All these parameters are estimated during training. If a specific parameter (or an entire layer's parameters) is not intended to have a patient-specific component, its corresponding entries in $\boldsymbol{\eta}_i$ are fixed at zero.

\subsection*{Loss Function and Optimization.}
The NME objective function is optimized per epoch. For our regression task with squared error loss, $l(Y_{ij}, \hat{Y}_{ij}) = \frac{1}{2}(Y_{ij} - \hat{Y}_{ij})^2$, and assuming a diagonal person-specific parameter covariance $\boldsymbol{\Sigma} = \tau^2\mathbf{I}$ (implying $\boldsymbol{\Sigma}^{-1} = (1/\tau^2)\mathbf{I}$), the objective function, adapted from Equation (1) of Wörtwein \emph{et al.} \cite{Wortwein2023}, is:
\begin{equation}
\mathcal{L}(\bar{\boldsymbol{\theta}}, \{\boldsymbol{\eta}_i\}) = \sum_{i=1}^{m} \sum_{j=1}^{n_i} \frac{1}{\sigma^2} \frac{1}{2}\left(Y_{ij} - \mu_{ij}^{\text{NME}}\right)^2 + \sum_{i=1}^{m} \boldsymbol{\eta}_i^{\top}\boldsymbol{\Sigma}^{-1}\boldsymbol{\eta}_i
\label{eq:NME_loss_combined} 
\end{equation}
where $\sigma^2$ is the observational (residual) variance, typically estimated as the mean squared error on the training data after each epoch. The first term encourages fidelity to the data, while the second term penalizes large deviations of person-specific parameters $\boldsymbol{\eta}_i$ from zero, effectively shrinking them towards the person-generic parameters $\bar{\boldsymbol{\theta}}$.

For stochastic gradient descent using mini-batches, the loss for a mini-batch $\mathcal{B}$ of size $B_{\text{batch}}$ (containing observations from a set of unique subjects $\mathcal{B}_{\text{subjects}}$) is formulated. The data fidelity part is the average loss over the batch. The regularization penalty is applied per subject within the batch, scaled by the proportion of that subject's total observations present in the current batch, as described by Wörtwein \emph{et al.} \cite{Wortwein2023}. Thus, the mini-batch loss is:
\begin{equation}
\tilde{\mathcal{L}}_{\mathcal{B}} = \frac{1}{B_{\text{batch}}} \sum_{(i,j) \in \mathcal{B}} \left[ \frac{1}{\sigma^2} \frac{1}{2}\left(Y_{ij} - \mu_{ij}^{\text{NME}}\right)^2 \right] + \sum_{k \in \mathcal{B}_{\text{subjects}}} \left( \frac{N_{k, \mathcal{B}}}{m_k} \boldsymbol{\eta}_k^{\top}\boldsymbol{\Sigma}^{-1}\boldsymbol{\eta}_k \right)
\label{eq:NME_minibatch_loss_wortwein_aligned} 
\end{equation}
where $N_{k, \mathcal{B}}$ is the number of observations for subject $k$ in the current mini-batch $\mathcal{B}$, and $m_k$ is the total number of training observations for subject $k$. This scaling ensures that the regularization for each subject is weighted according to its representation in the batch relative to its total contribution.

\subsection*{Gradient Updates.}
The parameters $(\bar{\boldsymbol{\theta}}, \{\boldsymbol{\eta}_i\}_{i=1}^m)$ are updated using gradients derived from the loss function $\mathcal{L}$ (Equation~\eqref{eq:NME_loss_combined}). For any parameter $\phi$ (which could be a component of $\bar{\boldsymbol{\theta}}$ or a component of some $\boldsymbol{\eta}_k$), the update uses its partial derivative.

Let $E_{ij} = Y_{ij} - \mu_{ij}^{\text{NME}}$ be the prediction error for subject $i$ at visit $j$. The derivative of the data fidelity part of the loss with respect to the model output $\mu_{ij}^{\text{NME}}$ (assuming squared error loss $l(Y, \hat{Y}) = \frac{1}{2}(Y-\hat{Y})^2$) is $\frac{\partial l}{\partial \mu_{ij}^{\text{NME}}} = -(Y_{ij} - \mu_{ij}^{\text{NME}}) = -E_{ij}$. Thus, the common error signal propagated back from the loss, scaled by the residual variance, is:
\[
\delta_{ij}^{\text{out}} = -\frac{E_{ij}}{\sigma^2}
\]
For our defined two-hidden-layer network, where $\mu_{ij}^{\text{NME}}$ is defined by equations \eqref{eq:NME_hidden1_activation_combined_final}, \eqref{eq:NME_hidden2_activation_combined_final}, and \eqref{eq:NME_output_two_hidden_combined_final}, we have $g_0'(x)=1$ (identity output activation), and $g_{1a}'(\cdot)$ and $g_{1b}'(\cdot)$ are the derivatives of the ReLU activation functions for the first and second hidden layers, respectively. Let $\boldsymbol{s}^{(1)}_{ij}$ and $\boldsymbol{s}^{(2)}_{ij}$ be the pre-activations for the first and second hidden layers.

The gradients for the weight parameters are derived as follows. These equations illustrate how the error signal is backpropagated and combined with local inputs/activations to update each parameter.

\vspace{\baselineskip}
\subsection*{Output Layer Parameters:}
The output layer directly computes the prediction $\mu_{ij}^{\text{NME}}$.
\noindent For an element $p$ of the \emph{generic} output weights $\bar{\boldsymbol{\omega}}^{(0)}$ (connecting $p$-th unit of the second hidden layer to the output):
\[
\frac{\partial\mathcal{L}}{\partial\bar{\omega}^{(0)}_p} = \sum_{i=1}^{m} \sum_{j=1}^{n_i} \delta_{ij}^{\text{out}} \cdot \alpha^{(2)}_{ij,p}
\]
This gradient term aggregates the product of the output error signal and the corresponding activation from the second hidden layer across all observations.
\noindent For the $p$-th element of a \emph{person-specific deviation} of an output weight $\boldsymbol{\eta}_{\omega^{(0)},k}$ for subject $k$:
\[
\frac{\partial\mathcal{L}}{\partial\eta_{\omega^{(0)}_p, k}} = \sum_{j=1}^{n_k} \delta_{kj}^{\text{out}} \cdot \alpha^{(2)}_{kj,p} + 2[\boldsymbol{\Sigma}^{-1}\boldsymbol{\eta}_k]_{\omega^{(0)}_p}
\]
where the second term serve as a regularization term that penalizes large deviations.

\vspace{\baselineskip}
\subsection*{Second Hidden Layer Parameters:}
Gradients for the second hidden layer involve backpropagating the error signal through the output layer weights.
\noindent For an element $\bar{\Omega}^{(2)}_{pc}$ of the \emph{generic} weights of the second hidden layer (connecting $c$-th unit of the first hidden layer to $p$-th unit of the second hidden layer):
\[
\frac{\partial\mathcal{L}}{\partial\bar{\Omega}^{(2)}_{pc}} = \sum_{i=1}^{m} \sum_{j=1}^{n_i} \delta_{ij}^{\text{out}} \cdot (\bar{\omega}^{(0)}_p + \eta_{\omega^{(0)}_p,i}) \cdot g_{1b}'(s^{(2)}_{ij,p}) \cdot \alpha^{(1)}_{ij,c}
\]
Here, the error signal is weighted by the effective output weight and the derivative of the second hidden layer's activation, then multiplied by the activation from the first hidden layer.
\noindent For an element $\eta_{\Omega^{(2)}_{pc},k}$ of a \emph{person-specific deviation} of a second hidden layer weight for subject $k$:
\[
\frac{\partial\mathcal{L}}{\partial\eta_{\Omega^{(2)}_{pc},k}} = \sum_{j=1}^{n_k} \delta_{kj}^{\text{out}} \cdot (\bar{\omega}^{(0)}_p + \eta_{\omega^{(0)}_p,k}) \cdot g_{1b}'(s^{(2)}_{kj,p}) \cdot \alpha^{(1)}_{kj,c} + 2[\boldsymbol{\Sigma}^{-1}\boldsymbol{\eta}_k]_{\Omega^{(2)}_{pc}}
\]

\vspace{\baselineskip}
\subsection*{First Hidden Layer Parameters:}
Gradients for the first hidden layer involve further backpropagation through the second hidden layer weights.
\noindent For an element $\bar{\Omega}^{(1)}_{cl}$ of the \emph{generic} weights of the first hidden layer (connecting $l$-th input feature to $c$-th unit of the first hidden layer):
\[
\frac{\partial\mathcal{L}}{\partial\bar{\Omega}^{(1)}_{cl}} = \sum_{i=1}^{m} \sum_{j=1}^{n_i} \left( \sum_{p=1}^{k_2} \delta_{ij}^{\text{out}} \cdot (\bar{\omega}^{(0)}_p + \eta_{\omega^{(0)}_p,i}) \cdot g_{1b}'(s^{(2)}_{ij,p}) \cdot (\bar{\Omega}^{(2)}_{pc} + \eta_{\Omega^{(2)}_{pc},i}) \right) g_{1a}'(s^{(1)}_{ij,c}) \cdot X_{ij,l}
\]
where $k_2$ is the number of units in the second hidden layer. 

\noindent For an element $\eta_{\Omega^{(1)}_{cl},k}$ of a \emph{person-specific deviation} of a first hidden layer weight for subject $k$:
\begin{align*}
\frac{\partial\mathcal{L}}{\partial\eta_{\Omega^{(1)}_{cl},k}} = &\sum_{j=1}^{n_k} \left( \sum_{p=1}^{k_2} \delta_{kj}^{\text{out}} \cdot (\bar{\omega}^{(0)}_p + \eta_{\omega^{(0)}_p,k}) \cdot g_{1b}'(s^{(2)}_{kj,p}) \cdot (\bar{\Omega}^{(2)}_{pc} + \eta_{\Omega^{(2)}_{pc},k}) \right) g_{1a}'(s^{(1)}_{kj,c}) \cdot X_{kj,l} \\
&+ 2[\boldsymbol{\Sigma}^{-1}\boldsymbol{\eta}_k]_{\Omega^{(1)}_{cl}}
\end{align*}
The pre-activations are $s^{(1)}_{ij,c}$ for the $c$-th unit of the first hidden layer, and $s^{(2)}_{ij,p}$ for the $p$-th unit of the second hidden layer for observation $(i,j)$. 

Gradients for all bias terms ($\bar{\boldsymbol{\delta}}^{(1)}, \boldsymbol{\eta}_{\delta^{(1)},i}, \bar{\boldsymbol{\delta}}^{(2)}, \boldsymbol{\eta}_{\delta^{(2)},i}, \bar{\delta}^{(0)}, \boldsymbol{\eta}_{\delta^{(0)},i}$) follow analogously by applying the chain rule, where the input to the bias is 1. During mini-batch optimization, these sums are taken over the observations $(i,j)$ in the current mini-batch $\mathcal{B}$, and the regularization term's gradient is applied only for subjects $k$ whose parameters $\boldsymbol{\eta}_k$ are being updated.

\subsection*{Summary of the Algorithm}

The Neural Mixed Effects (NME) model is trained using an iterative, optimization-based procedure, which involves employing stochastic gradient descent (e.g., Adam optimizer) within each epoch to update the person-generic parameters $\bar{\boldsymbol{\theta}}$ and the person-specific deviations $\{\boldsymbol{\eta}_i\}_{i=1}^m$. The variance components, namely the observational (residual) variance $\sigma^2$ and the covariance matrix of the person-specific parameters $\boldsymbol{\Sigma}$, are generally updated between epochs. For instance, $\sigma^2$ can be estimated based on the mean squared error from the training data using the current parameter estimates. The covariance matrix $\boldsymbol{\Sigma}$ is often assumed to be diagonal (e.g., $\boldsymbol{\Sigma} = \tau^2\mathbf{I}$) for scalability and is updated based on the sample covariance of the current person-specific deviations. 

This iterative training procedure allows the NME model to learn both the overall population trend (via $\bar{\boldsymbol{\theta}}$) and subject-specific nonlinear deviations (via $\boldsymbol{\eta}_i$) simultaneously. The person-specific deviations are regularized by their prior distribution, typically governed by the estimated covariance structure $\boldsymbol{\Sigma}$, which helps prevent overfitting and allows for robust estimation even for subjects with limited data.

The overall iterative process is outlined below:

\begin{algorithm}[H]
\caption{General Training Procedure for the Neural Mixed Effects (NME) Model}
\label{alg:nme_training_summary}
\begin{algorithmic}[1]
\Require Scaled training features $\mathbf{X}' = \{\mathbf{X}'_{ij}\}$, Scaled training targets $Y' = \{Y'_{ij}\}$, Subject indices for observations.
\Require Architectural choices (e.g., number of layers, units), Number of epochs $E$, Learning rate $\eta$, Batch size $B$.
\State Initialize person-generic parameters $\bar{\boldsymbol{\theta}}$ (e.g., Xavier initialization).
\State Initialize person-specific deviations $\{\boldsymbol{\eta}_i\}_{i=1}^m$ (e.g., to zeros or small random values).
\State Initialize covariance matrix $\boldsymbol{\Sigma}$ (e.g., as a scaled identity matrix).
\State Initialize residual variance $\sigma^2$ (e.g., to 1 or based on an initial pass over the data).
\State Initialize optimizer (e.g., Adam with learning rate $\eta$).

\For{epoch = 1 to $E$}
    \State Shuffle training data $(\mathbf{X}', Y')$.
    \For{each batch $b$ of size $B$}
        \State For each observation $(k,j)$ in batch $b$ (subject $k$, observation $j$):
        \State \quad Compute prediction $\hat{Y}'_{kj} = f(\mathbf{X}'_{kj}; \bar{\boldsymbol{\theta}} + \boldsymbol{\eta}_k)$.
        
        \State Compute mini-batch loss $\tilde{\mathcal{L}}_b$ (e.g., based on Eq.~\eqref{eq:NME_minibatch_loss_wortwein_aligned}), using current $\sigma^2, \boldsymbol{\Sigma}$.

        \State Compute gradients w.r.t. $\bar{\boldsymbol{\theta}}$ and relevant $\{\boldsymbol{\eta}_k\}$ for subjects in the batch: $\nabla_{\bar{\boldsymbol{\theta}}} \tilde{\mathcal{L}}_b$, $\nabla_{\boldsymbol{\eta}_k} \tilde{\mathcal{L}}_b$.

        \State Update $\bar{\boldsymbol{\theta}}$ and relevant $\{\boldsymbol{\eta}_k\}$ using the optimizer step.
    \EndFor
    
    \State Update $\sigma^2$ based on the average squared residuals over the full training set using current $\bar{\boldsymbol{\theta}}, \{\boldsymbol{\eta}_i\}$.
    \State Update $\boldsymbol{\Sigma}$ based on the sample covariance of the current person-specific deviations $\{\boldsymbol{\eta}_i\}_{i=1}^m$.
    
    \State Adjust learning rate or check for early stopping criteria if applicable.
\EndFor

\State \Return Learned parameters $\hat{\bar{\boldsymbol{\theta}}}$, $\{\hat{\boldsymbol{\eta}}_i\}$, $\hat{\boldsymbol{\Sigma}}$, $\hat{\sigma}^2$.
\end{algorithmic}
\end{algorithm}

\subsection*{Implementation and Results}
Our application of the Neural Mixed Effects (NME) model to predict Total UPDRS scores was based on the publicly available PyTorch implementation provided by Wörtwein \emph{et al.} \cite{Wortwein2023} and tailored for our case.

Input voice features and \texttt{test\_time} were standardized. We configured the NME model with a two-hidden-layer MLP (32 units in the first layer, 16 in the second, both using ReLU activations) as the base network, applying person-specific random effects ($\boldsymbol{\eta}_i$) to the output layer's bias. The model was trained for 4000 epochs using the Adam optimizer and a batch size of 512. During training, the observational variance $\sigma^2$ and a diagonal person-specific parameter covariance $\boldsymbol{\Sigma} = \tau^2\mathbf{I}$ were updated iteratively, consistent with the NME framework.

Predictive accuracy was measured  using mean
squared error (MSE) and mean absolute error (MAE). Table~\ref{tab:nme_results} summarizes the key performance metrics.
\begin{table}[H]
\centering
\caption{Performance of the NME Model on the Test Set for Total UPDRS Prediction.}
\label{tab:nme_results}
\begin{tabular}{lrr}
\toprule
Model                       & MSE & MAE \\
\midrule
NME-MLP  & 103.4075 & 8.1786   \\
\bottomrule
\end{tabular}
\end{table}


\section{Analysis of the Results}

To ensure a rigorous comparison, all predictive metrics reported in Table \ref{tab:gnmm_results} were calculated on the \textbf{original} \texttt{total\_UPDRS} scale. Predictions from models trained on transformed targets (log-transformed LMM/GAMM and standardized Neural Networks) were inverse-transformed to the clinical domain before error computation.

Table \ref{tab:gnmm_results} reveals a stark contrast in performance. The traditional statistical models (LMM and GAMM) achieved low error rates (MSE $6.56$ - $7.70$), corresponding to a Root Mean Squared Error (RMSE) of approximately $2.5$ points. This indicates these models successfully utilized the subject-specific random effects to interpolate the patient's individual trajectory.

In contrast, the neural architectures (GNMM, NME, ANN) yielded MSE values in the range of $96$ - $114$. These values are approximately equal to the global variance of the dataset, implying that the deep learning models failed to effectively learn the subject-specific deviations ($b_i$) from the limited sample of $N=42$ subjects. Instead of personalizing the predictions, the neural networks essentially reverted to predicting the population mean, resulting in errors roughly $4\times$ higher than the mixed-effects models. This validates our hypothesis that while neural models offer theoretical flexibility, they require significantly larger cluster counts to outperform efficient semi-parametric methods in longitudinal telemonitoring.

\begin{table}[H]
\centering
\caption{Predictive performance (MSE and MAE) of six models on the original UPDRS scale. Test set consists of each subject's last test time point (42 total).}
\begin{tabular}{lcc}
\toprule
\textbf{Model} & \textbf{MSE} & \textbf{MAE} \\
\midrule
LMM                  & $7.70$      &  $2.25$    \\
GAMM                 & $\mathbf{6.56}$      &  $\mathbf{2.00}$    \\
1‑layer GNMM         & $96.82$ & $6.96$ \\
2‑layer GNMM         & $106.09$ & $7.56$ \\
NME-MLP              & $103.41$ & $8.18$   \\
ANN (no random effect) & $114.20$ & $8.47$ \\
\bottomrule
\end{tabular}
\label{tab:gnmm_results_main}
\end{table}
\begin{figure}[H]
\centering
\includegraphics[width=0.9\linewidth]{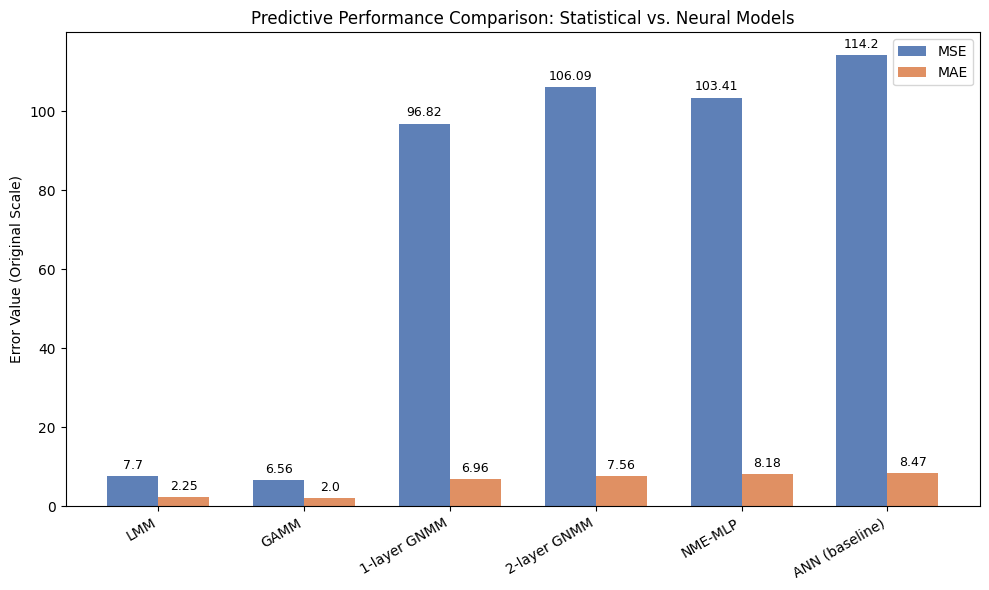}
\caption{Predictive performance comparison. The bar chart illustrates the magnitude of error (MSE and MAE) for each model, highlighting the superior performance of statistical methods (LMM, GAMM) over neural networks in this small-sample regime.}
\label{fig:model_comparison_bar}
\end{figure}

In this study, we developed and compared a suite of modeling approaches for predicting Parkinson’s disease progression, using a rich longitudinal voice dataset and the total UPDRS score as the clinical outcome. We evaluated both traditional statistical models, Linear Mixed Effects Models (LMM) and Generalized Additive Mixed Models (GAMM), as well as machine learning-based extensions, Generalized Neural Network Mixed Models (GNMM) and Neural Mixed Effect Models (NME-MLP).

To assess predictive performance, we constructed a test set composed of each subject’s last available time point (42 in total), while the remaining data were used for model training. This setup reflects a realistic clinical use case: forecasting future UPDRS values for already-observed patients, rather than for entirely new individuals.

Table \ref{tab:gnmm_results} reports the mean squared error (MSE) and mean absolute error (MAE) of each model. Among all methods, GAMM achieved the best performance with the lowest MSE ($6.56$) and MAE ($2.00$), indicating that the spline-based temporal effect captured meaningful nonlinear disease progression patterns. The LMM, although simpler, performed nearly as well (MSE = $7.70$), confirming the value of mixed-effects modeling with carefully selected covariates and interactions.

In contrast, the deep learning models, 1-layer and 2-layer GNMMs, ANN without random effects, and NME-MLP—performed substantially worse, with MSEs exceeding $96$ and MAEs exceeding 6, which is counter-intuitive. This finding is consistent with our recent evidence that larger, more complex biomedical deep models do not automatically improve clinical prediction over simpler baselines when the dataset and signal structure do not support the added capacity \cite{tong2025does}. As we normally assume that newer and complicated models outperforms the order and simpler ones. But that is not always the case, for any datasets which have many observations but only a modest number of predictors ($n \gg p$) a simple linear or spline-based model can already approximate the input–output mapping well, so the added capacity of deep networks does not translate into lower error unless it is strongly regularized.  While these architectures are expressive, their complexity and lack of explicit structure for within-subject correlation hinder their predictive accuracy in settings like ours.

As shown in Table \ref{tab:gnmm_results}, the GAMM significantly outperforms the neural baselines. To visualize this performance gap, we refer back to the trajectory complexity in Figure \ref{fig:three_figs}(c). The GAMM's spline components effectively capture these smooth, non-linear subject trends without the parameter bloat of the NME or GNMM. The neural models, by trying to learn these non-linearities from scratch with limited subjects, failed to converge to a generalizable solution, resulting in high variance across the five training runs.

This phenomenon underscores the critical role of \textbf{sample size efficiency} in clinical modeling. The LMM and GAMM impose structural assumptions (linearity and smoothness) that act as effective regularizers in small-cohort regimes ($N=42$). Conversely, the NME architecture is designed to optimize non-linear random effects, a task that requires a denser sampling of subject-specific trajectories to converge effectively. Therefore, the observed performance gap is not a failure of the neural architecture per se, but rather a demonstration that semi-parametric statistical methods provide a more robust and data-efficient baseline for current telemonitoring cohort sizes. While this study provides a rigorous benchmark on the Oxford dataset, the primary limitation is the cohort size ($N=42$). The poor performance of the NME model is directly attributable to the lack of sufficient subject-level clusters to map the manifold of random effects. Future work must focus on \textbf{external validation} using larger, multi-center datasets such as the Parkinson's Progression Markers Initiative (PPMI). Only with datasets exceeding hundreds of subjects can the theoretical flexibility of the NME framework be properly calibrated against the risk of overfitting.

\subsection*{The Role of Genetics, Lifestyle, and Medication}
While our study focused on acoustic biomarkers due to the constraints of the telemonitoring dataset, we acknowledge that PD progression is multifactorial. Recent literature highlights the significant impact of lifestyle factors and genetics on disease trajectories. For instance, Paul et al. \cite{Paul2021} demonstrated that lifestyle factors, including physical activity and diet, substantially modulate PD risk and progression. 

Genetic heterogeneity also plays a critical role in progression rates, which our current models absorb into the subject-specific random effects ($b_i$) rather than modeling explicitly. Studies have shown that patients with \textit{GBA} mutations often exhibit a "malignant" phenotype characterized by more rapid motor and cognitive decline compared to idiopathic PD \cite{Ortega2021, Simuni2020}. In contrast, \textit{LRRK2} carriers, particularly those with the G2019S mutation, often display a milder motor phenotype and slower progression of UPDRS scores \cite{SaundersPullman2018}. The omission of these genetic covariates likely inflates the unexplained variance ($\sigma^2$) in our mixed models, suggesting that future NME architectures should include static genetic embeddings to refine individual trajectory predictions.

Furthermore, the current dataset does not detail the ``On/Off'' medication status or specific levodopa equivalent daily doses (LEDD) for every recording. This is a critical limitation because therapeutic response introduces non-linear fluctuations in voice features that pure acoustic models may struggle to distinguish from disease progression. Systematic reviews indicate that while Levodopa significantly improves limb rigidity and bradykinesia, its effect on axial symptoms like speech is inconsistent; it may modify fundamental frequency ($F_0$) and jitter but often fails to improve vocal intensity or articulation \cite{Pinho2018}. This discrepancy means that a patient might show improvement in motor-UPDRS (limb scores) due to medication, while their voice biomarkers remain unchanged or degrade, confusing the model's mapping function. Finally, distinct clinical subtypes—such as the "diffuse malignant" versus "mild motor-predominant" phenotypes identified by Fereshtehnejad et al. \cite{Fereshtehnejad2017}—suggest that progression is not just rate-variable but structurally different across subgroups. Future telemonitoring frameworks should aim to integrate these multimodal data streams—combining voice, genetics, lifestyle factors, and pharmacodynamics—into the mixed-effects architecture.

\section{Summary and Future Work}

In summary, this study presented the first benchmarking of the Neural Mixed Effects (NME) framework against traditional statistical baselines for Parkinson's telemonitoring. Our results demonstrate that while neural architectures offer theoretical flexibility, classical semi-parametric models—specifically the GAMM—outperformed deep learning approaches (MSE 6.56 vs. 103.41) in the small-sample regime typical of clinical trials ($N=42$). Our findings indicate that incorporating smooth effects and subject-level random structures remains the most robust and interpretable strategy for the near-term forecasting of disease severity when high-volume data is unavailable.

However, our analysis of residuals and literature review (Section 4.1) identified critical sources of unmodeled variance that define the roadmap for future research. First, the impact of genetic heterogeneity (e.g., the rapid progression of \textit{GBA} carriers versus the slower trajectory of \textit{LRRK2} mutations) suggests that future NME architectures must move beyond pure time-series inputs. A promising direction is the development of \textbf{static-dynamic hybrid networks}, where static genetic embeddings and demographic factors are fused with dynamic voice vectors in the lower layers of the NME to inform the subject-specific parameter deviations ($\bm{\eta}_i$). Similarly, the confounding effect of Levodopa cycles requires the integration of pharmacokinetic pharmacodynamic (PK/PD) layers that can explicitly model the non-linear "On/Off" oscillations in voice features, distinct from the long-term neurodegenerative trend.

To further delineate the scalability of these architectures, future work will also focus on \textbf{external validation} using larger, multi-center datasets such as the Parkinson's Progression Markers Initiative (PPMI). While our current study establishes GAMM as the optimal choice for pilot cohorts, assessing NME on datasets exceeding hundreds of subjects will allow us to empirically map the sample size threshold where the theoretical flexibility of neural networks begins to overtake the data efficiency of statistical baselines.

Another key technical limitation of the current neural approaches is their lack of an explicit \emph{variable-selection} mechanism. Neither the GNMM nor the NME-MLP currently implements automated feature pruning, forcing reliance on trial-and-error which is inefficient for high-dimensional acoustic sets. Future work should focus on integrating sparsity-inducing penalties, such as the $\ell_{1}$ Group LASSO or Bayesian spike-and-slab priors, directly into the NME loss function. This would allow the model to automatically discard uninformative biomarkers, enhancing both generalization and clinical interpretability. While post-hoc explainability methods like SHAP or LRP have been applied to PD models \cite{Alshammari2025, Rezaei2024}, developing mixed-effects models with \textit{intrinsic} sparsity would provide more reliable trust mechanisms for clinicians. Building on the benchmark guidelines summarised by Tong \textit{et al.}~\cite{Tong2025}, we plan to construct a transparent test bed that rigorously compares these sparse neural extensions against classical methods.

Finally, the ultimate goal is to embed these models into a seamless telemedicine workflow \cite{yue2025revolutionizing}. However, clinical translation faces significant hurdles regarding standardization and privacy, particularly concerning device heterogeneity. In a real-world Bring-Your-Own-Device (BYOD) scenario, variations in smartphone microphones and background noise processing can introduce non-biological variance in Jitter and Shimmer features. To address this, future models must incorporate domain adaptation techniques or be trained on data augmented with simulated channel noise to ensure robustness across different recording environments.

In parallel with standardization efforts, privacy concerns dictate the need for secure deployment strategies. To comply with strict data privacy regulations such as HIPAA and GDPR, transmitting raw voice data to the cloud presents a substantial risk. A deployable system should therefore utilize edge computing, where feature extraction and the forward pass of the light-weight GAMM occur locally on the patient's device. In this architecture, only the computed risk scores—rather than the raw voice recordings—would be transmitted to the clinician's dashboard. Achieving this vision requires robust calibration procedures and clear data-governance protocols, and pilot studies integrating these components will be essential to demonstrate clinical utility before large-scale deployment.

\noindent\textbf{\textsc{SUPPORTING INFORMATION}}

Web Appendices referenced in Sections 3 and 4 are provided in the Supporting Information. Python and R code, along with a simulated example, are available at \url{https://github.com/RanTongUTD/Parkinson-Prediction/}.
\nocite{*} 

\end{document}